\documentclass[10pt]{article}

\usepackage[preprint]{tmlr}

\usepackage[utf8]{inputenc}
\usepackage[T1]{fontenc}
\usepackage{graphicx}
\usepackage{amsmath}
\usepackage{amssymb}
\usepackage{booktabs}
\usepackage{listings}
\usepackage{xcolor}
\usepackage{caption}
\usepackage{subcaption}
\usepackage{url}
\usepackage[hidelinks]{hyperref}
\usepackage{enumitem}
\usepackage{microtype}
\usepackage{wrapfig}

\makeatletter
\newif\ifanonymous
\if@accepted\anonymousfalse\else\anonymoustrue\fi
\def\@maketitle{\vbox{\hsize\textwidth
\if@preprint
  \centering
  {\LARGE\bf\sffamily \@title\par}\vskip 0.9em
  {\normalsize \@author\par}
\else
  {\LARGE\bf\sffamily \@title\par}\vskip \aftertitskip
  \if@accepted
    \@startauthor \@author \\ \\ {\bf Reviewed on OpenReview:} \openreview \@endauthor
  \else
    \@startauthor Anonymous authors\\Paper under double-blind review \@endauthor
  \fi
\fi
\vskip 0.22in}}
\makeatother

\definecolor{codebg}{RGB}{248,248,248}
\definecolor{codegreen}{RGB}{0,128,0}
\definecolor{codegray}{RGB}{128,128,128}
\definecolor{codeblue}{RGB}{0,0,180}
\setlist{nosep,leftmargin=*}

\graphicspath{{../../plots/}}

\title{From Language Models to Control Policies:\\Search, Selection, and Verification}

\author{%
Nima H. Siboni\textsuperscript{1,*}\qquad
Seyedreza Kiamousavi\textsuperscript{2}\qquad
Vahid Rostami\textsuperscript{3}\qquad
Emad Scharifi\textsuperscript{2}\\[0.7em]
{\small
\textsuperscript{1}Juna.ai, Berlin, Germany\\[0.15em]
\textsuperscript{2}Institute of Metal Forming, RWTH Aachen University, Aachen, Germany\\[0.15em]
\textsuperscript{3}Computational Systems Neuroscience, Institute of Zoology, University of Cologne, Cologne, Germany}\\[0.45em]
{\small\textsuperscript{*}Corresponding author: \texttt{nima@juna.ai}}%
}

\def\month{MM}
\def\year{YYYY}
\def\openreview{\url{https://openreview.net/forum?id=XXXX}}

\begin{document}
\maketitle

\begin{abstract}
  Industrial process control demands policies that are interpretable and auditable, requirements that black-box neural policies struggle to meet. LLM-driven heuristic synthesis offers an alternative: a language model iteratively proposes and refines human-readable Python controllers using rich behavioral feedback from a simulator. We study this form of programmatic policy search on hot steel rolling, a physics-simulated, multi-pass control problem with competing objectives and hard equipment constraints, and address three questions that arise once the search is run at scale: how to split a fixed budget of LLM iterations between restarting and refining, how to deploy the resulting pool of heuristics, and how to verify it. Treating each search run as a randomized algorithm, we find across 52 independent runs that runs of moderate length match the hindsight-optimal budget allocation, and that Luby-style universal restarts reach this regime without problem-specific tuning. Because each heuristic is cheap to simulate, a small portfolio selected per job outperforms the best single heuristic, also under cross-validation and on out-of-distribution operating conditions on which the best heuristic stalls. Finally, an automated audit pipeline combining static analysis, Z3 proofs, and property-based testing proves key safety properties for all 174 synthesized heuristics; the stall, which neither the search reward nor these properties reveal, shows that the specifications must also cover progress.
\end{abstract}

\section{Introduction}
\label{sec:intro}

Hot steel rolling is a multi-pass sequential decision problem: a heated slab is reduced in thickness over a series of passes, and at each pass the operator chooses a height reduction, an interpass waiting time, and a rolling velocity, balancing dimensional accuracy, metallurgical quality (grain size, temperature), equipment constraints, and process efficiency. Pass schedules are traditionally designed by experienced engineers using empirical rules and iterative simulation, which limits adaptability to new product specifications.

Reinforcement learning (RL) can learn effective policies for such problems, but the resulting neural networks are opaque: a domain expert cannot inspect \emph{why} a particular action was chosen, making deployment in safety-critical settings difficult. One-shot code generation by large language models (LLMs) offers interpretability but lacks the iterative refinement needed for complex multi-objective control.

Recent work on LLM-driven heuristic search~\citep{funsearch,eoh,reevo} has shown that iterative code generation can discover effective strategies in combinatorial domains. We apply this paradigm to industrial process control, where interpretability and safety auditability are first-class requirements, and view it as \emph{programmatic policy synthesis by LLM search}, with hot rolling as a physics-grounded, constrained, multi-objective benchmark. Once such a search is run at scale, three practical questions dominate: how to split a fixed budget of LLM iterations between restarting and refining, how to deploy a pool of synthesized heuristics rather than a single winner, and how to establish safety properties for the whole pool. We focus on these questions rather than on a comparison with neural RL policies, whose policy class and sample budgets differ and to which the formal audit does not apply. Our contributions are:
\begin{enumerate}
  \item A \textbf{budget allocation analysis across 52 independent runs} that treats the search as a randomized algorithm and shows that runs of moderate length match the hindsight-optimal allocation, and a \textbf{Luby-style universal restart} strategy that reaches this regime without retrospective analysis.
  \item \textbf{Deployment by simulator-selected portfolios}: because each heuristic is cheap to evaluate, a small portfolio of complementary heuristics can be run on each new job and the best result kept; the gain over the best single heuristic survives cross-validation and out-of-distribution operating conditions.
  \item A \textbf{five-layer automated audit pipeline} that checks each heuristic against domain-grounded safety, monotonicity, responsiveness, and consistency specifications, proving key safety properties with Z3 and testing the rest, applied to all 174 synthesized heuristics.
  \item \textbf{A reusable benchmark}: the Gymnasium environment with a pinned simulator stack, the search logs, the two heuristic pools with their evaluations, and all analysis scripts.
\end{enumerate}

\section{Method}

\paragraph{Problem setting.} We consider policy synthesis in a scenario-conditioned episodic control problem. The policy class is deterministic programs \texttt{heuristic(info, action\_mask)} $\rightarrow$ action; the objective is the mean episodic return over a set of operating scenarios; the only oracle is a simulator; and the unit of cost is one LLM iteration (ideation, implementation, evaluation). This is the setting of programmatic and interpretable RL~\citep{verma2018pirl,bastani2018viper,landajuela2021dsp} with the LLM as the proposal distribution, and it inherits the test-time-compute question of parallel sampling versus sequential refinement studied for LLM reasoning~\citep{brown2024monkeys,snell2024scaling}.

\subsection{Problem Formulation}

The environment, \textsc{FlatRollingEnv}, wraps the PyRolL physics simulator~\citep{pyroll} in a Gymnasium-compatible interface~\citep{gymnasium}. The state is a 10-dimensional vector comprising the current and target thickness, grain size, and temperature, plus rolling force, torque, height reduction (HR) limit, and step count. The action space has three components: height reduction (0--50\,mm in 0.1\,mm steps, 501~levels), interpass time (1--120\,s, 121~levels), and rolling velocity (7~discrete levels). Episodes terminate when the target thickness is reached or after 25~passes; height reductions below 1\,mm are executed as no-op passes.

Each candidate heuristic is a deterministic Python function:
\begin{lstlisting}[numbers=none,frame=none,backgroundcolor=\color{white}]
def heuristic(info: dict,
              action_mask: dict) -> list[int]
\end{lstlisting}
receiving the physical state and a validity mask that constrains actions to physically feasible values. Heuristics execute in a sandboxed environment restricted to \texttt{numpy} and \texttt{math}; before each step, the evaluator clips the returned actions into their index ranges and enforces the mask.

The reward function has five components: (1)~a step penalty ($-5.0$) encouraging efficiency, (2)~a grain size progress bonus with asymmetric penalties for undershooting, (3)~a height reduction efficiency bonus ($-5$ to 10) rewarding use of equipment capacity and penalising passes that exceed the force or torque limit, and terminal bonuses for (4)~grain size accuracy ($-2$ to 25) and (5)~temperature accuracy ($-2$ to 25). The search maximises the mean total reward over 8~\emph{search feedback scenarios} that vary thickness targets, grain size targets, equipment limits, and temperature targets (\S\ref{sec:setup}); the same scenarios provide the per-component feedback that guides refinement.

\subsection{Search Algorithm}

The search proceeds over $N$ outer iterations, each consisting of four phases (Figure~\ref{fig:convergence} shows the resulting reward trajectories across runs):

\textbf{Phase~1: Strategic Ideation.} The LLM generates $k{=}3$ distinct strategy proposals in natural language, each addressing reduction scheduling, interpass timing, velocity strategy, grain size management, and the roughing-to-finishing transition. This separation of strategic reasoning from code prevents premature commitment to implementation details.

\textbf{Phase~2: Idea Selection.} The LLM selects the most promising strategy with a justification grounded in the multi-objective reward structure.

\textbf{Phase~3: Implementation.} The LLM implements the selected strategy as executable Python via a tool-calling interface. Failed compilations or runtime crashes trigger up to 3~repair attempts.

\textbf{Phase~4: Evaluation and Feedback.} The heuristic is run on all 8~scenarios; for each, the system returns the total reward, step count, completion status, final state errors, per-component reward breakdowns, and action mask violations. This rich behavioral feedback, rather than a single scalar fitness, lets the LLM diagnose which reward components underperform and in which operating conditions.

\paragraph{Adaptive mode selection.} The framework employs three modes: \emph{Exploration} (no incumbent best---generate from scratch, informed by the top-$N$ best and last-$M$ recent heuristics), \emph{Refinement} (incumbent exists, stagnation below threshold---propose targeted single-variable modifications to the weakest reward component), and \emph{Radical Exploration} (stagnation exceeds threshold---demand fundamentally different strategies). When the conversation exceeds a length threshold, older messages are replaced by an LLM-generated summary that preserves strategies attempted, their results, key errors, and learnings; the most recent messages are retained verbatim.

\paragraph{LLM configuration.} We use Gemini~2.5~Pro~\citep{gemini} via the Google Generative AI API with function calling. The system prompt includes the complete environment and reward function source code alongside structured domain knowledge on S355 steel hot rolling metallurgy. Two tools are exposed: \texttt{read\_file} for inspecting source code and \texttt{suggest\_a\_heuristic} for submitting candidate implementations.

\section{Experimental Setup}
\label{sec:setup}

We conduct 52~independent search runs, totalling 730~iterations (mean 14.0 per run, range 1--30). The runs were made in groups over five days, and some settings were refined between groups (Appendix~\ref{app:configs}: Table~\ref{tab:config} summarises the settings and Table~\ref{tab:configs} lists every group); the evaluation scenarios, the reward, and the simulator are the same for all runs. Restricting the analyses to the 49 runs with Gemini 2.5 Pro, to the 27 runs with the final settings, or to the 17 runs that reached their planned length does not change the conclusions (Appendix~\ref{app:search_extra}).

The 8~search feedback scenarios change one aspect of a nominal scenario (100$\to$10\,mm, grain size 200$\to$12\,$\mu$m, temperature 1373$\to$1173\,K, HR limit 30\,mm) in two directions: (1)~thickness specifications of 90$\to$5\,mm and 110$\to$15\,mm, (2)~grain size specifications of 100$\to$5\,$\mu$m and 500$\to$25\,$\mu$m, (3)~HR limits of 20\,mm and 50\,mm, and (4)~temperature specifications of 1273$\to$1073\,K and 1523$\to$1273\,K. The completion rate (fraction of scenarios reaching target thickness) is a secondary search metric. For deployment evaluation, we additionally test on an \emph{in-distribution grid} of 81~scenarios that varies the same four axes one at a time (\S\ref{sec:inference}), and on 42~\emph{out-of-distribution} scenarios outside this grid (\S\ref{sec:shift}).

The 11~runs of 30~iterations took 36 to 73~minutes of wall time (median 40) and consumed a median of 5.7M~input and 129K~output tokens, whereas simulating one episode takes well under a second. All rewards are computed with \texttt{pyroll-core}~3.1.1 and \texttt{pyroll-integral-thermal}~3.0.0; because rewards depend on the simulator version, we release the full dependency lock file.

\section{Results}

\subsection{Baseline Heuristic}
\label{sec:baseline}

As a reference, a hand-coded baseline heuristic captures straightforward domain reasoning with three simple rules: (1)~\emph{height reduction}: apply 80\% of the per-pass HR limit until close to the target thickness, then use the remaining thickness to determine the final reduction; (2)~\emph{interpass time}: fixed at 10\,s regardless of conditions; (3)~\emph{rolling velocity}: a coarse six-level lookup on the temperature error (speed up when too cold, slow down when too hot), with no grain size or force awareness.

This baseline achieves a mean reward of 41.0 on the 8~search feedback scenarios (100\% completion) and 46.7 on the 81-case in-distribution grid. Its thickness precision is strong (0.04\,mm mean error), but it has no grain size control (7.8\,$\mu$m mean error---the final grain size is effectively constant regardless of target) and only coarse temperature management (38.8\,K mean error). It also incurs 4~constraint violations (3~force, 1~torque) across the 81~test cases.

By comparison, the best heuristic discovered by the LLM search achieves 55.5 on the 8~search scenarios and 60.9 on the in-distribution grid, 35\% and 31\% above the baseline. The improvement is concentrated in temperature control (7.7\,K vs.\ 38.8\,K mean error) and grain size management (6.3\,$\mu$m vs.\ 7.8\,$\mu$m), where the LLM-generated heuristic employs proportional controllers and strain-based scheduling that the baseline lacks. It does, however, incur more constraint violations: 13 (2~force, 11~torque) against the baseline's 4. The reward penalises passes that exceed these limits but does not forbid them, and the audit specifications (\S\ref{sec:audit}) concern the actions, not the resulting force and torque. This baseline is deliberately simple; a more carefully engineered controller could narrow the gap, so the comparison illustrates the reasoning the search discovers rather than a definitive advantage.

\subsection{Convergence Behavior}
\label{sec:convergence}

\begin{figure}[t]
\centering
\begin{minipage}[t]{0.48\linewidth}
\vspace{0pt}\centering
\includegraphics[width=\linewidth]{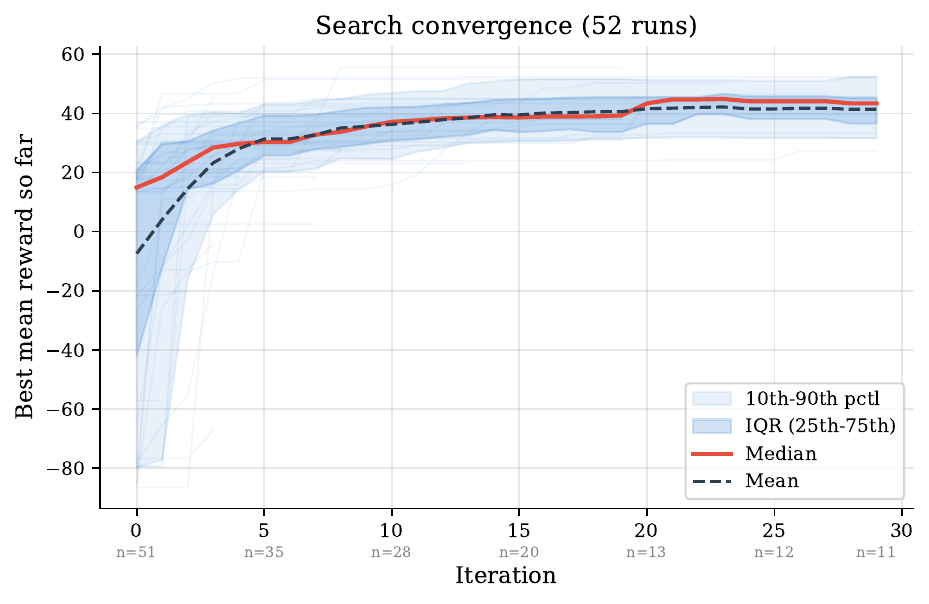}
\caption{Best-so-far reward across 52 independent runs: median, IQR, and 10th--90th percentile band, with individual traces in light blue. Labels give the number of runs contributing at each iteration.}
\label{fig:convergence}
\end{minipage}\hfill
\begin{minipage}[t]{0.48\linewidth}
\vspace{0pt}\centering
\includegraphics[width=\linewidth]{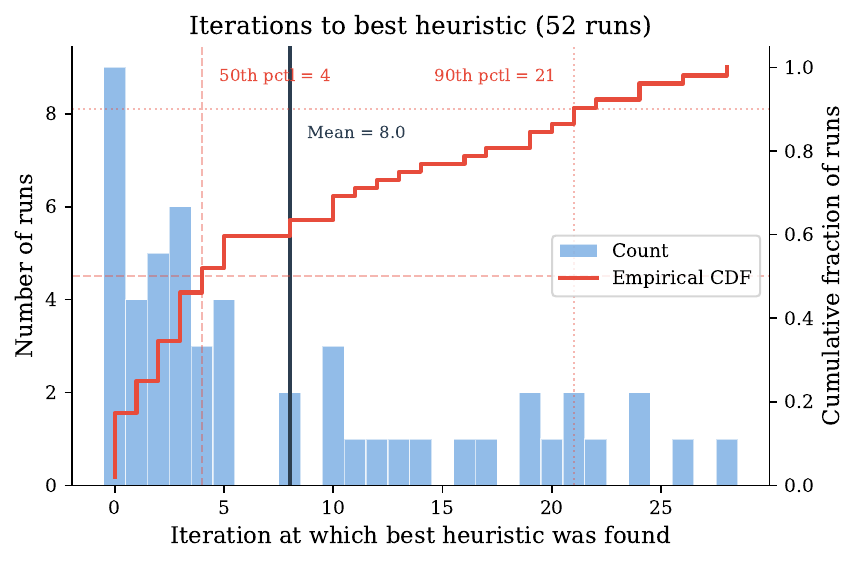}
\caption{Distribution of the iteration at which each run discovers its best heuristic. The 50th percentile is at iteration~4; the mean is at iteration~8.}
\label{fig:iters_to_best}
\end{minipage}
\end{figure}

The median best-so-far reward (Figure~\ref{fig:convergence}) rises to about 30 within the first 5~iterations and to about 40 by iteration~17, with diminishing returns thereafter; the interquartile range tightens over iterations, so most runs converge to a similar performance band despite exploring different strategies.

Figure~\ref{fig:iters_to_best} reveals that 50\% of runs find their best heuristic by iteration~4 (mean: iteration~8). This early convergence motivates the budget allocation analysis in \S\ref{sec:budget}.

The best single heuristic across all 52~runs achieves a mean reward of 55.5 and was discovered by only one run. It combines a strain-based height reduction schedule, a proportional temperature controller, and a force-adaptive rolling velocity (Listing~\ref{lst:heuristic}). We observe a strong tendency for the LLM to converge on domain-conventional strategies---in particular, Thermomechanical Controlled Processing (TMCP) paradigms---which we refer to as \emph{mode collapse to familiar architectures}.

\subsection{Budget Allocation}
\label{sec:budget}

A key practical question is how to allocate a fixed compute budget: run one long search or many short ones? We treat each search run as a randomized algorithm whose output after $L$ iterations, the best-so-far reward, is a random variable.
Comparing the expected marginal gain from continuing a run versus restarting fresh (Appendix~\ref{app:search_extra}, Figure~\ref{fig:continue_restart}), restarting yields higher marginal gain during the first few iterations, as each fresh start explores a new region of the strategy space; the two curves meet around iteration~4. This one-step comparison is myopic, however: it ignores that refinement compounds over later iterations, so we compare whole allocations.

\begin{wrapfigure}{r}{0.48\linewidth}
\centering
\includegraphics[width=\linewidth]{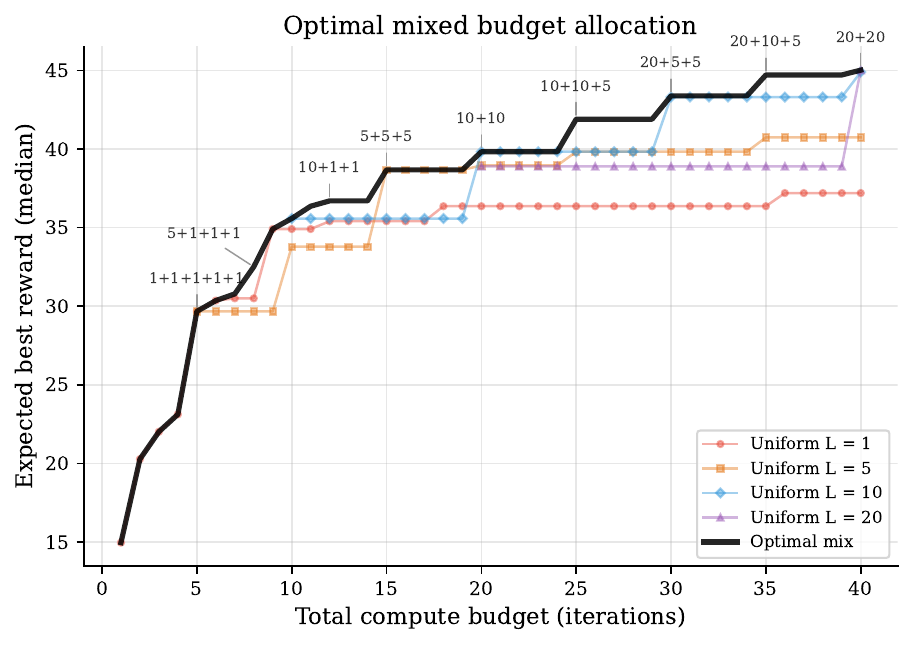}
\caption{Expected best reward (median) as a function of total iteration budget for uniform strategies ($L{=}1, 5, 10, 20$) and the optimal mixed allocation. Labels on the optimal mix curve show the allocation chosen.}
\label{fig:budget}
\end{wrapfigure}

Figure~\ref{fig:budget} formalises this with the empirical best-so-far distributions of the ad-hoc runs: for each run length~$L$, the best-so-far rewards after $L$ iterations of all runs that reached $L$ iterations define an empirical CDF~$F_L$. A budget of $T$ iterations allocated as independent runs of lengths $l_1, l_2, \ldots$ with $\sum_i l_i = T$ yields a global-best CDF of $\prod_i F_{l_i}$ (assuming independence); we report the median of this distribution. We compare uniform strategies (all runs of length $L \in \{1,5,10,20\}$) against the optimal mixed allocation.

By construction, the optimal mix is never worse than a uniform strategy, but its advantage is small. From a budget of 20~iterations on, the optimal allocations consist mostly of runs of 10 and 20~iterations (labels in Figure~\ref{fig:budget}), and uniform runs of 10 or 20~iterations come close to them whenever the budget is a multiple of the run length, whereas pure one-iteration sampling falls clearly behind as the budget grows. A bootstrap over runs (Appendix~\ref{app:search_extra}) shows that the remaining differences are within sampling noise: at $T{=}20$, for example, the optimal mix (10+10) is not reliably better than a single run of 20~iterations.

This finding---\emph{runs of moderate length (10--20 iterations) match the best allocation identified in hindsight, while one-iteration sampling and very long runs fall behind}---is a central practical recommendation: each run needs enough iterations to reach a reasonable local optimum, but not so many that budget is wasted on diminishing returns (runs of 30~iterations underperform at matched budget; Appendix~\ref{app:search_extra}). The sweet spot was identified in hindsight, however, which motivates a schedule that does not need it.

\subsection{Luby-Style Universal Restarts}
\label{sec:luby}

The budget allocation analysis above is retrospective: a practitioner starting a new problem cannot know the optimal mix in advance. Luby-style universal restarts~\citep{luby1993} offer a principled alternative: the restart schedule $[1,1,2,1,1,2,4,\ldots]$ (scaled by a unit~$u$) is provably within a logarithmic factor of the optimal fixed-cutoff strategy for \emph{any} unknown runtime distribution---requiring no problem-specific tuning. If the best fixed cutoff needs $\ell^*$ steps in expectation to succeed, the universal schedule needs $O(\ell^* \log \ell^*)$, and in the worst case no universal schedule can do better by more than a constant factor. The guarantee concerns the time until success, which we define as reaching a reward threshold (below).

We implement a Luby restart runner that executes 15~sub-runs with unit~$u{=}5$, yielding the schedule $[5, 5, 10, 5, 5, 10, 20, 5, 5, 10, 5, 5, 10, 20, 40]$ for a total of 160~iterations per campaign. Each sub-run is a fresh search with conversation reset (Gemini 2.5 Pro, temperature 1.0, three ideas, stagnation threshold~6, compaction memory; Appendix~\ref{app:configs}); optionally, the top~3 heuristics found so far are \emph{seeded} into later sub-runs as context that the LLM sees but need not refine. Table~\ref{tab:luby} compares two Luby campaigns with the 52 ad-hoc runs.

\begin{table}[t]
\setlength{\tabcolsep}{4pt}
\begin{minipage}[t]{0.48\linewidth}
\vspace{0pt}\centering
\caption{Luby restart campaigns vs.\ ad-hoc runs. Best: global best reward on the 8~search scenarios; Med.: median best per run; Iters: iterations that produced a heuristic (ad-hoc) or scheduled iterations (Luby); Wall: wall-clock time, summed over runs for the ad-hoc runs.}
\label{tab:luby}
\small
\begin{tabular}{@{}lccccr@{}}
\toprule
Strategy & Runs & Iters & Best & Med. & Wall \\
\midrule
Ad-hoc (52 runs) & 52 & 730 & 55.5 & 30.4 & ${\sim}$27\,h \\
Luby (unseeded) & 15 & 160 & 51.6 & 34.4 & 4.1\,h \\
Luby (seeded) & 15 & 160 & 48.3 & 43.2 & 4.1\,h \\
\bottomrule
\end{tabular}
\end{minipage}\hfill
\begin{minipage}[t]{0.48\linewidth}
\vspace{0pt}\centering
\caption{Expected iterations until the best-so-far reward reaches $\tau$: restarting every $L^*$ iterations (the best fixed cutoff; cost in parentheses) or every 30~iterations, and Luby schedules with unit~$u$ (mean of 20{,}000 simulated campaigns).}
\label{tab:threshold}
\small
\begin{tabular}{@{}rrrrrr@{}}
\toprule
 & & & \multicolumn{3}{c}{Luby} \\
\cmidrule(l){4-6}
$\tau$ & $L^*$ (cost) & $L{=}30$ & $u{=}1$ & $u{=}3$ & $u{=}5$ \\
\midrule
30 & 12 (6.2) & 9.0 & 8.4 & 8.0 & 7.7 \\
35 & 17 (12.4) & 19.2 & 15.8 & 15.2 & 14.2 \\
40 & 21 (20.2) & 33.7 & 45.1 & 32.9 & 36.2 \\
45 & 24 (40.5) & 61.0 & 101.2 & 82.9 & 68.9 \\
50 & 19 (76.0) & 140.0 & 214.1 & 138.1 & 110.8 \\
\bottomrule
\end{tabular}
\end{minipage}
\end{table}

\begin{figure}[t]
\centering
\begin{minipage}[t]{0.48\linewidth}
\vspace{0pt}\centering
\includegraphics[width=\linewidth]{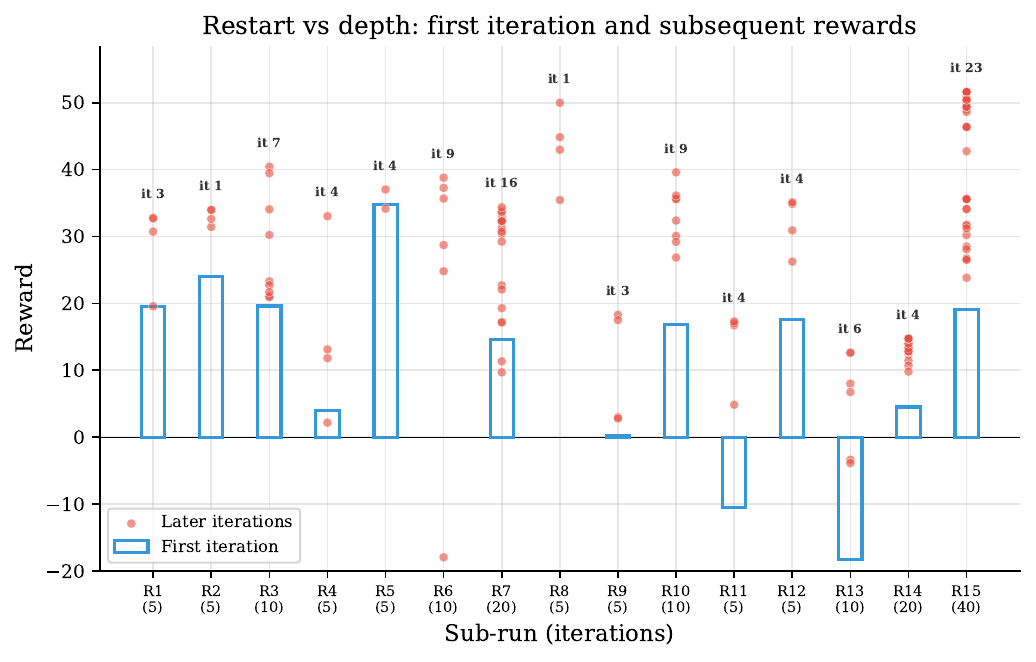}
\caption{Per-sub-run decomposition of an unseeded Luby campaign ($u{=}5$, 15~sub-runs, 160~iterations). Hollow bars: first-iteration reward (none where the first iteration produced no heuristic); dots: later iterations; labels: the iteration with each sub-run's best reward.}
\label{fig:luby_depth}
\end{minipage}\hfill
\begin{minipage}[t]{0.48\linewidth}
\vspace{0pt}\centering
\includegraphics[width=\linewidth]{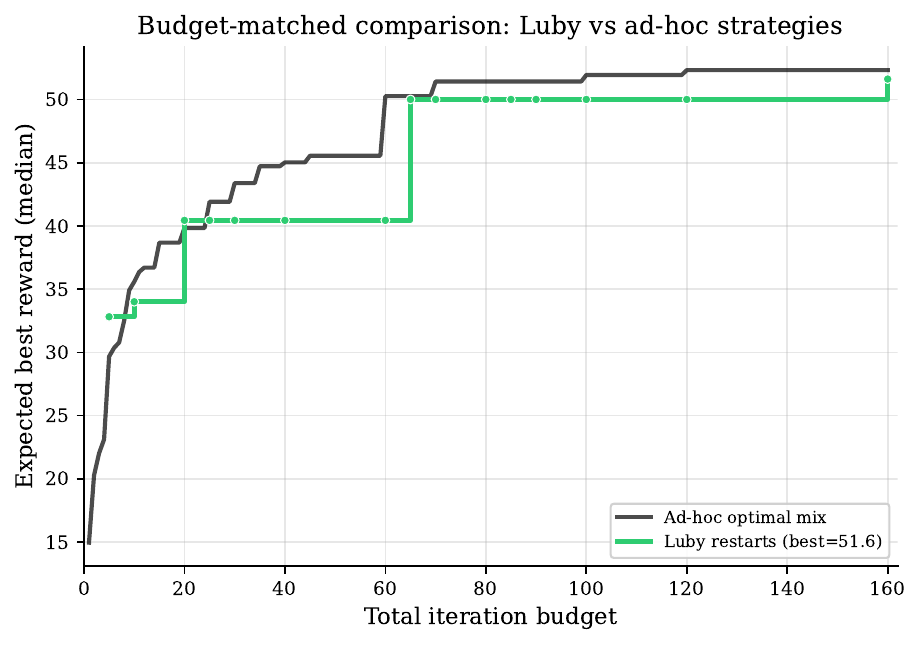}
\caption{Budget-matched comparison: expected best reward (median) of the optimal mixed allocation, computed in hindsight from 52~ad-hoc runs (730~iterations), and the cumulative best of a single unseeded Luby campaign (160~iterations).}
\label{fig:luby_budget}
\end{minipage}
\end{figure}

Figure~\ref{fig:luby_depth} separates the reward of each sub-run's first iteration (the ``restart'' contribution) from the rewards of its later iterations (the ``depth'' contribution). In 14~of~15 sub-runs, the best reward comes from a later heuristic than the first---typically from iterations 3 to 9---demonstrating that depth within each sub-run is essential. The exception is sub-run~8, whose first heuristic (from its second iteration; the first produced none) scored 50.0, illustrating the complementary value of fresh restarts: a lucky initialisation can outperform extended refinement.

\paragraph{Budget-matched comparison.} Figure~\ref{fig:luby_budget} compares the Luby campaign's actual cumulative best-so-far against the optimal mix curve, which took 52~ad-hoc runs totalling 730~iterations to estimate. With only 160~iterations in a single automated session, the Luby trajectory approaches the optimal mix, which required $4.6{\times}$ as many iterations and hindsight to identify. A single campaign is also a single sample, so we place it in context by simulating 20{,}000 Luby campaigns whose sub-runs are drawn from the empirical per-length distributions~$F_L$ (Appendix~\ref{app:search_extra}). The observed campaign lies at the 39th percentile of the simulated ones: typical, not lucky. On average, the simulated Luby schedule trails the optimal allocation by at most 1.3 reward points at every budget up to 160~iterations, and overtakes uniform runs of 5~iterations from about 10~iterations on and uniform runs of 30~iterations from about 50~iterations on.

\paragraph{Time to a reward threshold.} To apply Luby's guarantee literally, we define success as a best-so-far reward of at least~$\tau$. From the runs we estimate the expected number of iterations that restarting every $L$ iterations needs to succeed, $\mathbb{E}[\min(X, L)]/P(X \leq L)$, where $X$ is the number of iterations a single run needs to reach~$\tau$, and we simulate the Luby schedule on the same empirical hit times (Table~\ref{tab:threshold}; Appendix~\ref{app:search_extra}). The optimal cutoff depends on the target (12 to 24 iterations) and is unknown a priori. With $u{=}5$, Luby needs 1.1 to 1.8 times the iterations of the optimal cutoff for every target; it needs fewer iterations than restarting every 30~iterations for three of the five targets and at most 13\% more for the other two. With $u{=}1$, most of the budget goes to one-iteration runs and the cost reaches 2.8 times the optimum. This is the logarithmic-factor guarantee at work on an LLM search.

\subsection{Inference Strategies: From Pool to Deployment}
\label{sec:inference}

Multiple search runs produce a pool of diverse heuristics. Because each is a lightweight Python function---simulating an episode takes well under a second and auditing a heuristic a median of 0.05~seconds (\S\ref{sec:audit})---it is practical to evaluate the \emph{entire pool} on each new job specification, filter out candidates with audit failures, and keep the best result; the simulator serves as ground truth, so no learned classifier is needed.

\begin{wrapfigure}{R}{0.45\linewidth}
\centering
\includegraphics[width=\linewidth]{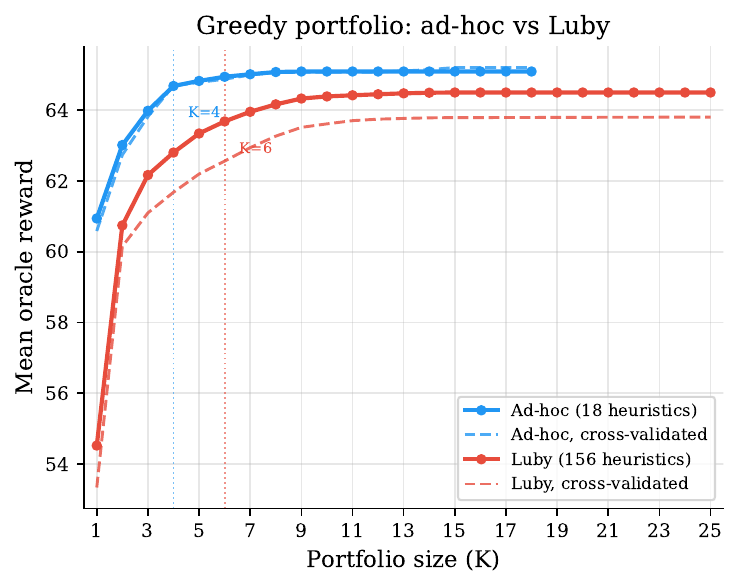}
\caption{Greedy portfolio construction: ad-hoc (18~heuristics) vs.\ Luby (156~heuristics), evaluated on the 81-case in-distribution grid. Solid: portfolios selected and scored on all 81~cases; dashed: selected on a random half of the cases and scored on the other half (mean over 200 splits). Dotted lines mark the $K$ at which 90\% of the full-pool gain is captured.}
\label{fig:portfolio}
\end{wrapfigure}

We apply greedy portfolio construction to both the ad-hoc pool and the Luby pool (Figure~\ref{fig:portfolio}): starting from an empty set, at each step $K$ we add the heuristic that maximises the \emph{oracle reward}---defined as the mean, over all test cases, of the maximum reward achieved by any portfolio member on that test case. This selects \emph{complementary} specialists: the second heuristic chosen is the one that best covers the cases where the first underperforms, not the second best on average. The ad-hoc pool contains the 18~heuristics of the run that produced the global best heuristic (20~iterations), the Luby pool the 156~heuristics of all 15~sub-runs of the unseeded campaign; both are evaluated on the 81-case in-distribution grid.

For the ad-hoc pool, the single best heuristic achieves 60.9 and the full oracle reaches 65.1---a portfolio of $K{=}4$ captures 90\% of the gain. The Luby pool starts lower (single best 54.5) because its heuristics are drawn from many short sub-runs, but it converges rapidly: by $K{=}5$ it reaches 63.3 (88\% of its 10.0-point gain), and the full oracle of 64.5 is within 1\% of the ad-hoc oracle, whose pool is the best of 52~runs (730~iterations), identified in hindsight. Even if evaluating many candidates per job is undesirable, a pruned portfolio of 3--5 heuristics provides near-oracle performance at minimal cost.

\paragraph{Cross-validation.} To check that the gain is not an artifact of selecting and scoring on the same 81~cases, we select portfolios on a random half of the cases and score them on the other half (200 splits; dashed in Figure~\ref{fig:portfolio}), and on three of the four variation axes and score them on the held-out axis. The gain survives. With random halves, the cross-validated curves stay within 1.2~points of the in-sample ones, and a four-heuristic portfolio scores 64.7 (ad-hoc) and 61.7 (Luby) on the unseen half, against 60.6 and 53.3 for the single best heuristic. Holding out a whole axis lowers all values but preserves the gap (62.4 and 59.1 against 58.1 and 48.0, averaged over the four axes).

\paragraph{Evaluation sets and reported metrics.} The reward of 55.5 (best heuristic across all 52~runs) is measured on the 8~search scenarios, whereas 60.9 (best single heuristic) and 65.1 (oracle portfolio) are measured on the 81-case grid, which was never used during search; its one-axis sweeps include easier scenarios, so the two sets are not directly comparable.

Although each run tends to collapse to familiar architectures (\S\ref{sec:convergence}), the heuristics it produces along the way, and those of different restarts, specialise on complementary subsets of the operating space.

\subsection{Robustness Outside the Search Distribution}
\label{sec:shift}

The in-distribution grid varies one search axis at a time around a nominal scenario. To test deployment beyond it, we evaluate all 174~synthesized heuristics and the baseline on 42~scenarios in eight families that leave these ranges, use off-grid combinations, or shift several axes at once (thicker slabs, thinner and thicker targets, tighter and looser mills, off-grid temperature and grain size pairs, and combined shifts; Appendix~\ref{app:shift}). The best single heuristic and the $K{=}4$ portfolio of each pool are selected on the 81-case grid and deployed unchanged; the portfolio again keeps the best member per case by simulation. Five episodes in which heuristics of one Luby sub-run raise an exception at an HR limit of 60\,mm, outside the audited ranges (Table~\ref{tab:input_ranges}), are excluded rather than counted as failures. Table~\ref{tab:shift} summarises the results; three findings stand out.

\begin{table}[t]
\centering
\caption{Mean episode return over the 42~out-of-distribution cases (completed cases in parentheses). Best single and $K{=}4$ portfolio are selected on the in-distribution grid; the oracle keeps the best heuristic of the pool per case; Median is the median heuristic. The hand-coded baseline scores 37.2 and completes 40 cases.}
\label{tab:shift}
\small
\setlength{\tabcolsep}{4pt}
\begin{tabular}{@{}lrrrr@{}}
\toprule
Pool & Best single & $K{=}4$ & Oracle & Median \\
\midrule
Ad-hoc (18) & 41.8 (37) & 58.7 (41) & 63.6 (42) & 39.1 \\
Luby (156) & 51.2 (42) & 59.9 (42) & 67.3 (42) & 36.1 \\
\bottomrule
\end{tabular}
\end{table}

\emph{In-distribution rank predicts shifted performance.} Across the 156 Luby-pool heuristics, the Spearman correlation between the mean on the 81-case grid and the mean on the 42 shifted cases is 0.94 (0.67 for the 18 ad-hoc heuristics), so selecting heuristics on the search distribution is a sound deployment rule.

\emph{The best single heuristic is brittle; the portfolios are not.} The best heuristic found by the search (reward 55.5) fails to terminate on 5 of the 42 shifted cases (three with thin targets and two combined shifts), and on these two families it scores below the hand-coded baseline (Appendix~\ref{app:shift}). The mechanism is instructive.
Rolling a 100\,mm slab to a 2\,mm target, the heuristic reaches a thickness of 5.400015\,mm, while the mask allows reductions of up to 3.4\,mm; the finishing branch (Listing~\ref{lst:heuristic}, lines~2--4) is therefore skipped by a floating-point margin, the strain branch requests a 0.5\,mm pass, which the environment executes as a no-op, and the same state repeats until the episode ends. None of the 11~audit specifications can see this: the heuristic never overshoots, but the catalogue contains no progress property. A one-line specification---if at least 1\,mm remains and the mask allows a 1\,mm pass, then HR $\geq$ 1\,mm---would have flagged it. The four-heuristic portfolio of the same pool fails on only one case, and that of the Luby pool completes all~42.

\emph{Observation noise lowers returns but causes no failed episodes.} With multiplicative Gaussian noise of up to 5\% on the observed thickness, grain size, temperature, force, and torque, the selected heuristics and portfolios complete every episode, and at 5\% noise the portfolios stay 3.5 (Luby) and 5.7 (ad-hoc) points above the respective best single heuristic (Appendix~\ref{app:shift}).

\subsection{Interpretability of the Search Process}
\label{sec:search_interp}

Unlike black-box optimization, the LLM search produces a complete reasoning trace: at every iteration, the model articulates its strategy in natural language \emph{before} generating code. Table~\ref{tab:search_trace} (Appendix~\ref{app:trace}) shows a representative excerpt from a 20-iteration run, illustrating how the LLM's understanding deepens through interaction with the evaluation feedback.

Three patterns emerge: the LLM explicitly diagnoses weaknesses from the per-component feedback (iteration~1), discovers trade-offs between objectives (iteration~9) and eventually synthesizes solutions that balance them (iteration~14), and its radical departures (iteration~10) regress before enabling later breakthroughs---the hybrid at iteration~14 combines insights from both the conservative and radical branches.

\subsection{Interpretability of the Generated Policy}

Each generated policy is a readable Python function with explicit control logic. Listing~\ref{lst:heuristic} shows a condensed excerpt from the best-performing heuristic (reward~55.5). The code reveals three layers of metallurgical reasoning:

\begin{enumerate}
  \item \textbf{Hybrid height reduction} (\texttt{lines~2--18}): A finishing pass removes the remaining thickness whenever the mask allows it; otherwise a strain-based target with grain-size-aware scheduling applies. The ``conservative'' branch (lines~7--9) is dead code: any positive remaining thickness within the mask has already returned at line~4.
  \item \textbf{Proportional temperature control} (\texttt{lines~21--26}): Interpass time computed as a proportional controller on temperature error, with a 25\,K deadband and linear scaling up to 40\,s of cooling.
  \item \textbf{Force-aware velocity} (\texttt{lines~29--35}): Rolling velocity adapts to force as a fraction of the equipment limit, reducing speed when force exceeds 70\% capacity.
\end{enumerate}

A domain expert can inspect each decision, verify it against metallurgical principles, and modify specific thresholds before deployment, which a neural network policy does not allow. Manual inspection does not scale, however: a search run produces dozens of candidates, each with effectively infinitely many input combinations. This motivates the automated audit described next.

\begin{figure}[t]
\begin{minipage}[t]{0.475\linewidth}
\begin{lstlisting}
# --- Finishing Pass ---
remaining = current_thickness - target_thickness
if 0 < remaining <= max_hr_mm:
    return [round(remaining * 10), 1, 3]

# --- Height Reduction Logic ---
if remaining < max_hr_mm:
    # CONSERVATIVE: split into two passes
    target_hr = remaining * 0.5
else:
    # AGGRESSIVE: strain-based reduction
    if grain_size > target_grain * 2.0:
        strain = 0.8   # heavy refinement
    elif grain_size > target_grain:
        strain = 0.4   # moderate refinement
    else:
        strain = 0.1   # preservation
    target_hr = thickness * (1 - exp(-strain))
\end{lstlisting}
\end{minipage}\hfill
\begin{minipage}[t]{0.475\linewidth}
\begin{lstlisting}[firstnumber=20]
# --- Interpass Time (Temperature Control) ---
temp_error = temperature - target_temperature
if temp_error > 25.0:
    frac = clip((temp_error - 25) / 125, 0, 1)
    interpass = 5.0 + frac * 35.0  # 5-40s
else:
    interpass = 1  # minimal wait

# --- Rolling Velocity (Force Management) ---
force_limit = 4_000_000
if force > force_limit * 0.70:
    velocity = 2     # slow for safety
elif force < force_limit * 0.25:
    velocity = 6     # fast for throughput
else:
    velocity = 5     # default
\end{lstlisting}
\end{minipage}
\captionof{lstlisting}{Condensed excerpt from the best heuristic (reward 55.5). Helper functions and variable unpacking omitted; \texttt{max\_hr\_mm} is the largest height reduction the action mask allows.}
\label{lst:heuristic}
\end{figure}

\subsection{Automated Safety Auditing}
\label{sec:audit}

Deploying LLM-generated code in safety-critical industrial settings requires more than interpretability: the code must be systematically verified against domain specifications. Inspired by the formalized specification approach of SpecVerify~\citep{specverify}, we developed a five-layer automated audit pipeline that verifies each candidate heuristic:
%
\begin{enumerate}
  \item \textbf{AST structural \& security checks} (29~checks, 8~categories): validates function signature, sandbox compliance, action mask usage, bounds clipping, division safety, and return path completeness.
  \item \textbf{Interval analysis}: propagates input value ranges through the heuristic's arithmetic to statically verify that all three outputs remain within action bounds.
  \item \textbf{Domain specifications}: 11~formal requirements (SPEC-001--011) covering safety (no overshoot, HR-limit respect), monotonicity, responsiveness, and consistency properties.
  \item \textbf{Z3 SMT verification}: proves specs hold for \emph{all} valid inputs (the ranges of Table~\ref{tab:input_ranges}) by translating the heuristic AST to Z3 constraints. Safety specs use a scalar decomposition of the mask, and monotonicity specs compare two symbolic copies of the heuristic (two-translation technique). The proofs are of partial correctness: they cover the arithmetic and control flow that the translator supports and hold for executions that return normally; outputs that depend on anything else, as well as exceptions, array indexing, and output bounds, are left to property-based testing.
  \item \textbf{Property-based testing}: randomized testing (207~inputs including 7~edge cases) verifies specs that Z3 cannot handle due to complex numpy operations, and catches runtime exceptions.
\end{enumerate}


As a worked example, Table~\ref{tab:audit} shows the audit of the best heuristic (reward~55.5): 65~checks with zero errors. Z3 proves the three safety properties (no overshoot, HR-limit respect, non-negative reduction) for all valid inputs. Because the heuristic reads the mask array and calls helper functions, which the translator cannot follow, the remaining specifications are deferred to property-based testing and pass with zero failures, although the tests miss a violation of SPEC-004 (Appendix~\ref{app:population}).

The audit takes a median of 0.05~seconds per heuristic (at most 0.24~seconds), so it can also run inside the search loop as a filter that rejects unsafe candidates before they enter the pool.

\begin{table}[t]
\centering
\caption{Audit results for the best heuristic (reward~55.5). The pipeline produces 65~checks with 0~errors. Z3 proves safety specifications for \emph{all} valid inputs; remaining specs verified via randomized property testing.}
\label{tab:audit}
\small
\begin{tabular}{@{}lrcl@{}}
\toprule
Layer & Checks & Pass & Notes \\
\midrule
AST (8 categories) & 29 & 27/29 & 2 warn \\
Interval analysis & 9 & 8/9 & 1 warn \\
Z3 SMT prover & 10 & 10/10 & 3 proved $\forall$, 7 deferred \\
Property testing & 17 & 17/17 & 207 inputs, 0 failures \\
\midrule
\multicolumn{2}{@{}l}{Total: 65 checks} & 62/65 & \textbf{0 errors} \\
\bottomrule
\end{tabular}
\end{table}

\begin{figure}[t]
\begin{minipage}[t]{0.46\linewidth}\vspace{0pt}\centering
\includegraphics[width=\linewidth]{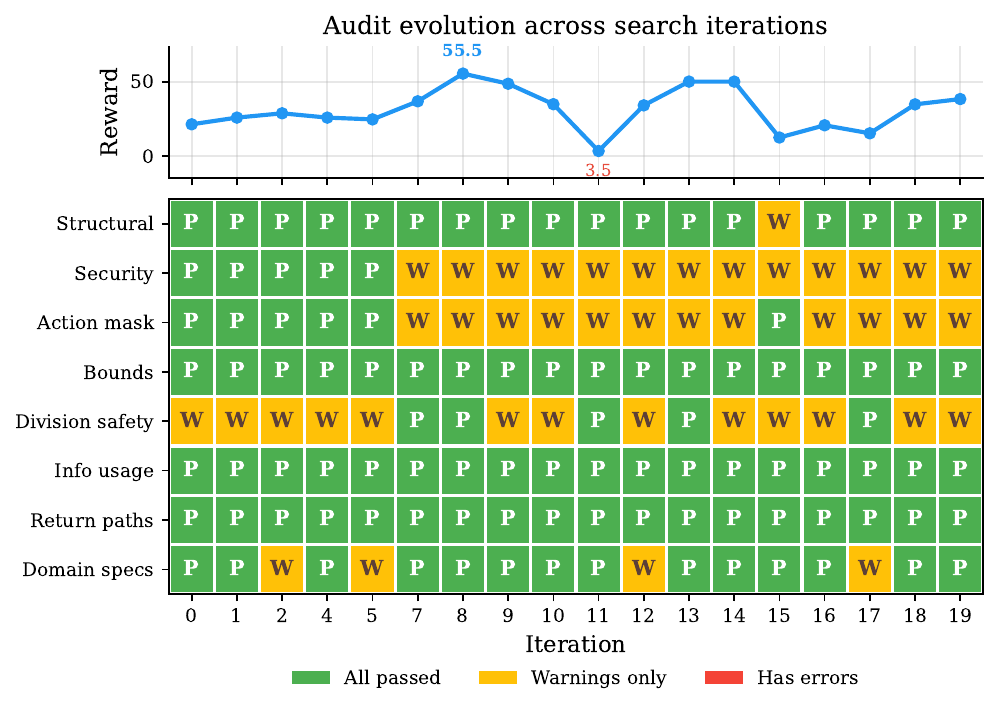}
\caption{Traffic-light heatmap of per-category audit results for the 18~heuristics of the best ad-hoc run (the same run used for portfolio construction in \S\ref{sec:inference}; iterations 3 and 6 produced none). Green = all checks passed, amber = warnings only, red = errors.}
\label{fig:audit_evolution}
\end{minipage}\hfill
\begin{minipage}[t]{0.525\linewidth}\vspace{0pt}\centering
\captionsetup{type=table}\renewenvironment{table}[1][]{}{}\let\small\footnotesize
\begin{table}[t]
\centering
\caption{Population audit of all 174~synthesized heuristics (156 from the Luby pool and 18 from the ad-hoc pool). Z3 columns: number of heuristics for which the spec was proved for all valid inputs, refuted with a counterexample (cex), or deferred to property-based testing; the last column counts heuristics violating the spec on the 207 test inputs (including heuristics that raise exceptions).}
\label{tab:audit_population}
\small
\setlength{\tabcolsep}{4pt}
\begin{tabular}{@{}llrrrr@{}}
\toprule
 & & \multicolumn{3}{c}{Z3} & Prop. \\
\cmidrule(lr){3-5}
Spec & Property & proved & cex & deferred & fail \\
\midrule
001 & no overshoot & 174 & 0 & 0 & 0 \\
002 & HR-limit respect & 174 & 0 & 0 & 0 \\
003 & HR $\geq 0$ & 174 & 0 & 0 & 0 \\
004 & HR mono.\ in thickness & 0 & 0 & 174 & 13 \\
005 & wait mono.\ in grain & 98 & 0 & 76 & 0 \\
006 & velocity mono.\ in force & 45 & 2 & 127 & 0 \\
007 & HR responsive & 0 & 0 & 174 & 0 \\
008 & wait responsive & 0 & 0 & 174 & 26 \\
009 & velocity responsive & 0 & 0 & 174 & 92 \\
010 & deterministic & 0 & 0 & 0 & 0 \\
011 & continuous & 0 & 0 & 174 & 2 \\
\midrule
\multicolumn{6}{@{}l}{No error-level finding: 174/174 (100\%); median audit time 0.05\,s} \\
\bottomrule
\end{tabular}
\end{table}

\end{minipage}
\end{figure}

\paragraph{Population audit.} A single example does not show what the pipeline finds in practice, so we run it on all 174~heuristics of the two pools (Table~\ref{tab:audit_population}; Appendix~\ref{app:population}). Z3 proves the three safety specifications (SPEC-001--003) for every one of them and the monotonicity specifications SPEC-005 and SPEC-006 for 98 and 45, with 2~counterexamples; it defers SPEC-004 for all 174, because every heuristic derives its height reduction from the mask array. None has an error-level finding. Running the pipeline at this scale did expose defects in the audit tool itself, which we fixed before producing Table~\ref{tab:audit_population} (Appendix~\ref{app:population}): action masks that differed from the environment's made 22~heuristics appear to violate the safety specifications or the output bounds, the interval analyser crashed on 2~heuristics, and the Z3 translator kept stale values for code it could not translate, which let it prove SPEC-004 for 63~heuristics, the best one included, on a constant height reduction. Verification tools for generated code need tests of their own. Warning-level findings correlate weakly and \emph{positively} with reward (Spearman $\rho = 0.16$): better heuristics are longer and trip more static checks, so warning counts are not a quality signal. The responsiveness specifications also diagnose what the search did not learn: 53\% of the heuristics never change rolling velocity with force or temperature on the test inputs. Finally, the catalogue contains no progress property; \S\ref{sec:shift} shows a failure that only such a property would catch.

Figure~\ref{fig:audit_evolution} shows how audit results evolve across the 18~heuristics of the best ad-hoc run. Bounds, information usage, and return paths pass in every iteration---these categories are effectively enforced by the sandbox and prompt constraints. From iteration~7 on, the heuristics define helper functions inside the policy, whose names the allowed-names check (SEC-005) does not recognise, and all except iteration~15 lack an explicit fallback for masked actions (MSK-003); iteration~15 keeps controller state in a global variable (structural). Division safety and domain specification compliance show intermittent warnings, but no iteration produces an error. Within this run, warning counts do not track reward: the best iteration~8 (reward~55.5) has 3~warnings, while iteration~15 (reward~12.6) has 7.

\section{Discussion}

\paragraph{What works.} The structured 4-phase loop with rich behavioral feedback is effective: the LLM generates competent policies within a few iterations, and these embed recognizable metallurgical reasoning (grain-size-aware strain scheduling, proportional temperature control, force-adaptive velocity), although this knowledge reaches the LLM only through the system prompt. Interpretable outputs and an audit with formal guarantees (\S\ref{sec:audit}) are a qualitative advantage over RL in safety-critical settings.

\paragraph{What fails.} Four limitations emerged. First, \emph{radical exploration is unreliable}: three of the four radical departures in one 30-iteration run produced architectures that crashed catastrophically (rewards $-80$ to $-89$), suggesting that the current ``try something completely different'' prompt lacks sufficient guidance. Second, \emph{conversation compaction causes amnesia}: effective micro-modifications (e.g., strain-aware roughing, proportional temperature control) were rediscovered several times within a run, so compaction fails to preserve actionable insights (Appendix~\ref{app:memory} compares memory strategies). Third, \emph{mode collapse to familiar architectures} limits diversity within a run: the LLM gravitates toward TMCP-style controllers and rarely discovers unconventional mappings. Fourth, \emph{safety is not progress}: the best heuristic never overshoots and passes every safety specification, yet stalls on out-of-distribution targets because of a floating-point edge case that no specification in the catalogue covers (\S\ref{sec:shift}).

\paragraph{Practical recommendations.} Based on our analysis: (1)~allocate budget to runs of 10--20~iterations rather than to one-shot sampling or very long runs---or use Luby-style restarts with a unit of about 5~iterations (\S\ref{sec:luby}) to reach this regime without problem-specific tuning, (2)~deploy a small portfolio of 3--5 complementary heuristics selected by simulation rather than the single best heuristic, which can be brittle out of distribution (\S\ref{sec:shift}), (3)~invest in structured memory mechanisms to mitigate conversation amnesia, and (4)~run the automated audit, which takes a median of 0.05~seconds per heuristic, on all deployment candidates and extend its catalogue with progress specifications (\S\ref{sec:audit}).

\paragraph{Scope.} All results come from one environment and one LLM. The restart analysis rests on one corpus of ad-hoc runs and a single unseeded Luby campaign, which we place in context by simulation rather than by repetition; because runs were stopped at different lengths, and more often when they were doing poorly, the per-length distributions are estimated from different subsets of runs. The conclusions hold on the runs that reached their planned length and on the runs with the final settings (Appendix~\ref{app:search_extra}). We treat the rich behavioral feedback as a design choice and do not ablate it, and we do not compare against neural RL policies (\S\ref{sec:intro}).

\section{Related Work}

\textbf{LLM-driven heuristic and code generation.} FunSearch~\citep{funsearch} couples LLMs with an evolutionary algorithm to discover mathematical functions at a scale of ${\sim}$1M evaluations, Evolution of Heuristics (EoH)~\citep{eoh} evolves (thought, code) pairs with the LLM as crossover and mutation operator, and ReEvo~\citep{reevo} adds reflection with separate generator and critic roles. Concurrent work by \citet{evogineer} evolves policies for LunarLander by population-based search with behavioral feedback; we use a single-trajectory search with adaptive modes on a multi-objective industrial domain. MLES~\citep{mles} adds \emph{visual} feedback from rollout videos, whereas we use per-component reward decompositions, and Vulcan~\citep{vulcan} applies LLM-driven heuristic search to systems problems such as cache eviction and memory tiering.

\textbf{Programmatic and interpretable policies.} Representing policies as programs is an established route to interpretable and verifiable RL. PIRL~\citep{verma2018pirl} searches a policy language guided by a neural oracle, VIPER~\citep{bastani2018viper} extracts decision-tree policies that can be formally verified, LEAPS~\citep{trivedi2021leaps} searches a learned latent space of programs, and deep symbolic policy search~\citep{landajuela2021dsp} discovers compact symbolic control laws; genetic programming has produced interpretable policies for an industrial benchmark~\citep{hein2018gp}. LLM program search replaces these hand-designed proposal mechanisms with a language model: Code as Policies~\citep{liang2023cap} generates robot policy code directly, Eureka~\citep{ma2024eureka} evolves reward code, LLM-guided search has synthesized interpretable controllers for classical control tasks~\citep{bosio2025synthesizing}, and AlphaEvolve~\citep{novikov2025alphaevolve} and LLaMEA~\citep{vanstein2024llamea} evolve algorithms. The questions we study---budgeting, deploying, and auditing the resulting pool---apply to any of these search procedures.

\textbf{Inference compute and restarts.} How to spend a fixed budget of LLM calls has been studied for reasoning as parallel sampling versus sequential refinement: coverage keeps growing with the number of independent samples~\citep{brown2024monkeys}, and the compute-optimal mix of sampling and revision depends on problem difficulty~\citep{snell2024scaling}. For randomized search, the classical answer is restarts: \citet{luby1993} give a universal schedule within a logarithmic factor of the optimal fixed cutoff for any run-time distribution, randomized restarts eliminate the heavy tails of combinatorial search~\citep{gomes1998boosting}, and Luby's schedule performed best in aggregate among the restart policies compared for a clause-learning SAT solver~\citep{huang2007restarts}. We apply this analysis to LLM-driven policy synthesis, where one iteration costs minutes and success is defined by a reward threshold.

\textbf{Hot rolling optimization.} Traditional approaches use genetic algorithms~\citep{ga_rolling}, mathematical programming, or expert systems for pass schedule design. Recent work applies RL to simplified rolling models~\citep{rl_rolling}, but typically with analytical rather than physics-based simulators. Our environment uses PyRolL~\citep{pyroll} for physically grounded state transitions including realistic force/torque calculations and microstructural evolution.

\section{Conclusion}

We studied LLM-driven heuristic synthesis for hot steel rolling as programmatic policy search, demonstrating that this paradigm can produce auditable control policies in a realistic industrial simulator. Across 52~independent runs, runs of moderate length match the hindsight-optimal budget allocation, and Luby-style universal restarts~\citep{luby1993} reach this regime without prior data or manual tuning. Because synthesized heuristics are cheap to evaluate, a small portfolio selected by simulation outperforms the best single heuristic under cross-validation and out of distribution, where the best single heuristic stalls on a floating-point edge case. The automated audit formally proves the three safety specifications for all 174~synthesized heuristics; the stall shows that its catalogue also needs progress specifications. Future work includes structured architectural mutations to counteract mode collapse, persistent cross-run memory, Z3 support for array-dependent specifications, the audit as an in-loop filter, and adaptive schedules that adjust the Luby unit online. \ifanonymous Code, search logs, heuristic pools, and analysis scripts are provided as supplementary material.\else Code and data are available at \citet{pso-repo}.\fi

\bibliography{refs}
\bibliographystyle{tmlr}

\appendix
\raggedbottom
\section{Domain Specification Catalog}
\label{app:specs}

Table~\ref{tab:specs} lists all 11~domain specifications verified by the audit pipeline. Each specification encodes a physical requirement of the hot rolling process as a formal property that can be checked automatically. Specifications are grouped into four categories reflecting different aspects of controller correctness.

\textbf{Safety} specifications (SPEC-001--003) are rated \textsc{error} severity: violation means the controller could damage equipment or produce out-of-spec steel. No-overshoot (SPEC-001) ensures the height reduction never thins the slab below the target gauge---a fundamental constraint in rolling, since under-thick slabs cannot be corrected. HR-limit respect (SPEC-002) prevents exceeding the mill's per-pass reduction capacity.

\textbf{Monotonicity} specifications (SPEC-004--006) encode expected physical relationships. SPEC-004 requires that a thicker slab (larger remaining reduction) triggers a proportionally larger height reduction---a controller that reduces \emph{less} when the slab is further from target would be pathological. SPEC-005 requires longer interpass waiting times when grain size is below target, since grain growth during the interpass period is the primary mechanism for achieving target microstructure. SPEC-006 requires that higher rolling force produces lower velocity, protecting the mill from overload.

\textbf{Responsiveness} specifications (SPEC-007--009) verify that each output dimension actually responds to its relevant inputs. A controller that outputs a constant height reduction regardless of thickness (failing SPEC-007) would be degenerate. These are existential properties: we check that there \emph{exist} input conditions producing different outputs.

\textbf{Consistency} specifications (SPEC-010--011) verify determinism and continuity. SPEC-010 ensures no hidden stochasticity (which would make the controller unpredictable). SPEC-011 checks that small input perturbations do not cause discontinuous output jumps---important for smooth mill operation.

\begin{table}[h]
\centering
\caption{Complete domain specification catalog. Severity: E\,=\,Error (safety-critical), W\,=\,Warning, I\,=\,Info. Method: Z3\,=\,formally proved; H\,=\,property testing.}
\label{tab:specs}
\small
\begin{tabular}{@{}llp{6cm}cl@{}}
\toprule
ID & Category & Property & Sev. & Meth. \\
\midrule
001 & Safety & HR $\leq$ thickness $-$ target & E & Z3 \\
002 & Safety & HR/10 $\leq$ \texttt{hr\_limit} & E & Z3 \\
003 & Safety & HR $\geq 0$ & E & Z3 \\
\midrule
004 & Monoton. & Thicker $\Rightarrow$ larger HR & W & H \\
005 & Monoton. & Smaller grain $\Rightarrow$ longer wait & I & H \\
006 & Monoton. & Higher force $\Rightarrow$ lower velocity & I & H \\
\midrule
007 & Respons. & HR varies with thickness & W & H \\
008 & Respons. & Interpass varies with temp/grain & W & H \\
009 & Respons. & Velocity varies with force/temp & I & H \\
\midrule
010 & Consist. & Deterministic output & E & H \\
011 & Consist. & Continuous output & W & H \\
\bottomrule
\end{tabular}
\end{table}

The ``Method'' column shows the verification technique for the best heuristic (reward~55.5): Z3 proves the three safety properties (SPEC-001--003); the monotonicity specifications are deferred to property-based testing because the heuristic's outputs depend on helper functions and the mask array, which the translator cannot follow (Appendix~\ref{app:z3}), and the responsiveness and continuity specifications are always tested. Table~\ref{tab:audit_population} reports the outcome of every specification for all 174~synthesized heuristics.

The catalogue contains no progress (liveness) specification. Section~\ref{sec:shift} describes a provably safe heuristic that stalls because it requests reductions below the 1\,mm no-op threshold; a progress specification such as ``if at least 1\,mm remains and the mask allows a 1\,mm pass, then HR $\geq$ 1\,mm'' would flag this and can be checked with the same scalar mask decomposition as SPEC-001--003.

\section{Verification Methodology Details}
\label{app:verification}

\subsection{AST-Based Static Analysis (Layer 1)}
\label{app:ast}

The first audit layer performs 29~static checks organized into 8~categories, executed via AST visitors without running the heuristic. Table~\ref{tab:ast_checks} summarizes the categories.

\begin{table}[h]
\centering
\caption{AST-based check categories. Each category targets a specific failure mode of LLM-generated code.}
\label{tab:ast_checks}
\small
\begin{tabular}{@{}lrp{8cm}@{}}
\toprule
Category & Chk & What it catches \\
\midrule
Structural integrity & 4 & Wrong signature, missing returns, global state, class definitions \\
Security / sandbox & 6 & Imports, exec/eval, file I/O, dunder access, forbidden modules \\
Action mask compliance & 3 & Mask not referenced, not checked, no fallback \\
Bounds \& clipping & 3 & Outputs outside $[0,500]$, $[1,120]$, $[1,6]$ \\
Division safety & 2 & Unguarded divisions, NaN/Inf risk \\
Info dict usage & 4 & Missing critical keys, force unawareness \\
Return path complete. & 3 & Dead paths, wrong length, no fallback \\
Control logic quality & 4 & Ignores thickness, constant outputs \\
\bottomrule
\end{tabular}
\end{table}

Security checks deserve special attention because LLMs occasionally generate code that stores mutable state on the function object (e.g., \texttt{heuristic.prev\_hr = val}) or accesses dunder attributes. The audit flags these via SEC-006 (allowed modules check), which requires all attribute-set targets to be whitelisted. In the best ad-hoc run (Figure~\ref{fig:audit_evolution}), no iteration stores state on the function object or accesses dunder attributes, but iteration~15 keeps the state of a PID controller in a global variable, which the structural check STR-003 flags; the security warnings from iteration~7 on come from SEC-005, which does not recognise names defined in helper functions nested inside the heuristic. Other runs in our 52-run corpus do exhibit these patterns, confirming the check's value.

\subsection{Interval Analysis (Layer 2)}
\label{app:intervals}

The interval analysis layer propagates input value ranges through the heuristic's arithmetic to verify output bounds \emph{statically}---without executing the code. Starting from known input ranges (Table~\ref{tab:input_ranges}), the evaluator walks the AST and computes a closed interval $[\text{lo}, \text{hi}]$ for every intermediate variable. For example, given \texttt{current\_thickness} $\in [5, 110]$ and \texttt{target\_thickness} $\in [5, 15]$, the expression \texttt{remaining\_reduction = current\_thickness - target\_thickness} yields the interval $[-10, 105]$.

\begin{table}[h]
\centering
\caption{Input value ranges for interval analysis and Z3 constraints.}
\label{tab:input_ranges}
\small
\begin{tabular}{@{}lrrl@{}}
\toprule
Variable & Min & Max & Unit \\
\midrule
current\_thickness & 5.0 & 110.0 & mm \\
target\_thickness & 5.0 & 15.0 & mm \\
hr\_limit & 20.0 & 50.0 & mm \\
stock\_temperature & 800 & 1523 & K \\
target\_temperature & 1073 & 1273 & K \\
current\_grain\_size & 5 & 500 & $\mu$m \\
target\_grain\_size & 5 & 25 & $\mu$m \\
rolling\_force & $-$100 & $4{\times}10^6$ & N \\
rolling\_torque & $-$100 & $1.3{\times}10^5$ & N$\cdot$m \\
step\_count & 0 & 25 & --- \\
\bottomrule
\end{tabular}
\end{table}

Interval arithmetic handles all standard operations: addition (intervals add), subtraction (lo subtracts hi), multiplication (product of all endpoint pairs), and division (with explicit zero-crossing detection). The evaluator also handles \texttt{np.clip}, \texttt{min}, \texttt{max}, \texttt{int()}, \texttt{round()}, and if/else branching (union of branch intervals). The output is three interval checks (RNG-001--003) verifying that the HR, interpass, and velocity outputs remain within their action bounds for all possible input combinations. Bounds that the analysis cannot limit stay unbounded through \texttt{int()}, \texttt{round()}, \texttt{floor()}, and \texttt{ceil()}, and the corresponding output check is then reported as a warning.

\subsection{Z3 Formal Verification (Layer 4)}
\label{app:z3}

The Z3 verification layer translates the heuristic's Python AST into Z3 SMT constraints and proves that domain specifications hold for \emph{all} valid inputs---not just sampled ones. The proofs establish partial correctness: they cover the arithmetic and control flow that the translator supports and hold for executions that return normally. Exception-freedom, array indexing, and output bounds are covered by property-based testing (Layer~5).

\paragraph{AST-to-Z3 translation.}
The translator (\texttt{HeuristicZ3Translator}) creates a symbolic Z3 variable for each of the 10~input keys (Table~\ref{tab:input_ranges}), with domain constraints bounding their ranges. It then walks the heuristic's function body, translating each Python construct:
\begin{itemize}
  \item \textbf{Arithmetic}: $+, -, \times, \div$, floor division, power $\rightarrow$ Z3 arithmetic
  \item \textbf{Comparisons}: $<, \leq, >, \geq, =, \neq$ $\rightarrow$ Z3 boolean expressions
  \item \textbf{Control flow}: \texttt{if/else} $\rightarrow$ Z3 \texttt{If(cond, then, else)} with branch merging; \texttt{return} statements are combined in source order under their path conditions
  \item \textbf{Calls}: \texttt{np.clip(x,lo,hi)} $\rightarrow$ \texttt{If(x<lo, lo, If(x>hi, hi, x))}; similarly for \texttt{max}, \texttt{min}, \texttt{abs}, \texttt{int}, \texttt{round}, \texttt{math.ceil/floor}
  \item \textbf{Dict access}: \texttt{info['key']} $\rightarrow$ corresponding Z3 variable
\end{itemize}
Constructs that cannot be translated---notably numpy array operations such as \texttt{np.where}, \texttt{np.argmin}, array indexing, and calls of helper functions---cause a \texttt{Z3TranslationError}. The translator then forgets rather than guesses: a failed assignment makes its targets unknown, an \texttt{if} statement whose condition cannot be translated makes the variables assigned in either branch unknown and, if it contains a \texttt{return}, every return unreliable, and a return value that references an unknown variable marks that output dimension as \texttt{None}, so that the spec is deferred to property-based testing.

\paragraph{Contiguous mask decomposition.}
The action mask for height reduction is a 501-element binary array where $\texttt{mask}[i] = 1$ iff $i/10 \leq \min(\texttt{hr\_limit},\, 0.7 \times \texttt{thickness},\, \texttt{thickness} - \texttt{target})$. Because this mask is contiguous (all 1s up to some index, then all 0s), it can be decomposed to a single scalar \texttt{max\_valid\_idx}. This avoids reasoning about arrays entirely and enables Z3 to prove safety properties (SPEC-001--003) by showing that any action within the mask satisfies the constraint.

\paragraph{Two-translation technique.}
Monotonicity and relational specifications require comparing the heuristic's output under two different input conditions. The verifier creates two independent translator instances with prefixed variable names (e.g., \texttt{a\_current\_thickness}, \texttt{b\_current\_thickness}), translates the heuristic AST \emph{twice}, and builds a property comparing the two outputs:
\begin{enumerate}
  \item Create translators $T_a$ (prefix \texttt{a\_}) and $T_b$ (prefix \texttt{b\_}).
  \item Translate the function body with each, extracting return expressions.
  \item Equalize all shared inputs: $\forall k \neq k_{\text{vary}}:\ a_k = b_k$.
  \item Add ordering constraint on the varied input: $a_{k_{\text{vary}}} > b_{k_{\text{vary}}}$.
  \item Prove the output relationship: e.g., $\texttt{out}_a \geq \texttt{out}_b$ for increasing monotonicity.
\end{enumerate}
Z3 then checks whether the \emph{negation} of this property is satisfiable. If unsatisfiable (UNSAT), the property is proved for all valid inputs. If satisfiable (SAT), Z3 provides a concrete counterexample. For SPEC-005 and SPEC-006, this technique proves or refutes the property for 98 and 47 of the 174 heuristics of Table~\ref{tab:audit_population}, each in well under a second, and defers the rest. For SPEC-004 (HR monotonicity) it defers every heuristic, because all of them derive the height reduction from the mask array (\texttt{np.where}, \texttt{np.argmin}, or a list comprehension), which is untranslatable.

\paragraph{Unknown values.}
A false proof arises whenever the translator reasons about a value that is not the one the code computes: one branch of an if/else translates but the other does not (e.g., \texttt{if remaining\_reduction < 0.1: hr\_action = 0} succeeds, but the else branch uses \texttt{np.argmin}), a reassignment fails and an earlier value of the variable remains, or an \texttt{if} statement with an untranslatable condition is skipped together with the \texttt{return} inside it. The population audit exposed the last two cases (Appendix~\ref{app:population}). The translator therefore keeps a variable only when it knows its value on every path that reaches the current statement: a variable defined in only one branch, or whose assignment failed, is removed from the environment, and a skipped return makes every output untranslatable. Tests of these cases are part of the code.

\subsection{Property-Based Testing (Layer 5)}
\label{app:properties}

The final layer executes the heuristic with actual numpy operations on 207~test inputs (200~random + 7~deterministic edge cases) and verifies all specs that Z3 could not handle.

\paragraph{Edge cases.}
Seven edge cases encode domain expertise about which operating conditions are most likely to trigger failures:
\begin{enumerate}
  \item \textbf{Near target}: thickness $= 10.1$\,mm, target $= 10.0$\,mm (tests minimal-reduction logic)
  \item \textbf{Maximum state}: all inputs at upper bounds (tests saturation behavior)
  \item \textbf{Minimum state}: all inputs at lower bounds (tests boundary handling)
  \item \textbf{Tight mask}: remaining reduction $= 0.5$\,mm (only a few valid HR actions)
  \item \textbf{Force sentinel}: force $= -100$\,N (pre-first-pass sentinel value from environment)
  \item \textbf{Equal grain sizes}: current $=$ target (tests zero-error handling)
  \item \textbf{Equal temperatures}: stock $=$ target (tests zero-error handling)
\end{enumerate}

\paragraph{Random inputs.}
The 200~random inputs are sampled uniformly from the ranges in Table~\ref{tab:input_ranges}, with the constraint that \texttt{current\_thickness} $\geq$ \texttt{target\_thickness}. For each input, a consistent action mask is generated from the physical constraints (HR limit, 70\% thickness rule, minimum gauge); the interpass and velocity masks have the environment's lengths (121 and 7 entries).

\paragraph{Per-input checks.}
Each test input runs 6~basic checks (PBT-001--006): no exceptions, correct return type, HR/interpass/velocity within bounds, and mask compliance (the selected HR action must not be masked out). Additionally, all domain specs deferred from Z3 are verified by calling the spec's check function with the test input.

\paragraph{Z3-to-Hypothesis deferral.}
The Z3 layer produces a list of deferred spec IDs that it could not verify. These are passed to the property testing layer, which verifies them via execution. The deferral is transparent in the audit report: each deferred spec appears in both the Z3 section (marked ``deferred to Hypothesis'') and the property testing section (marked ``deferred from Z3''), ensuring complete traceability.

\subsection{Population Audit}
\label{app:population}

\begin{figure}[t]
\centering
\includegraphics[width=\linewidth]{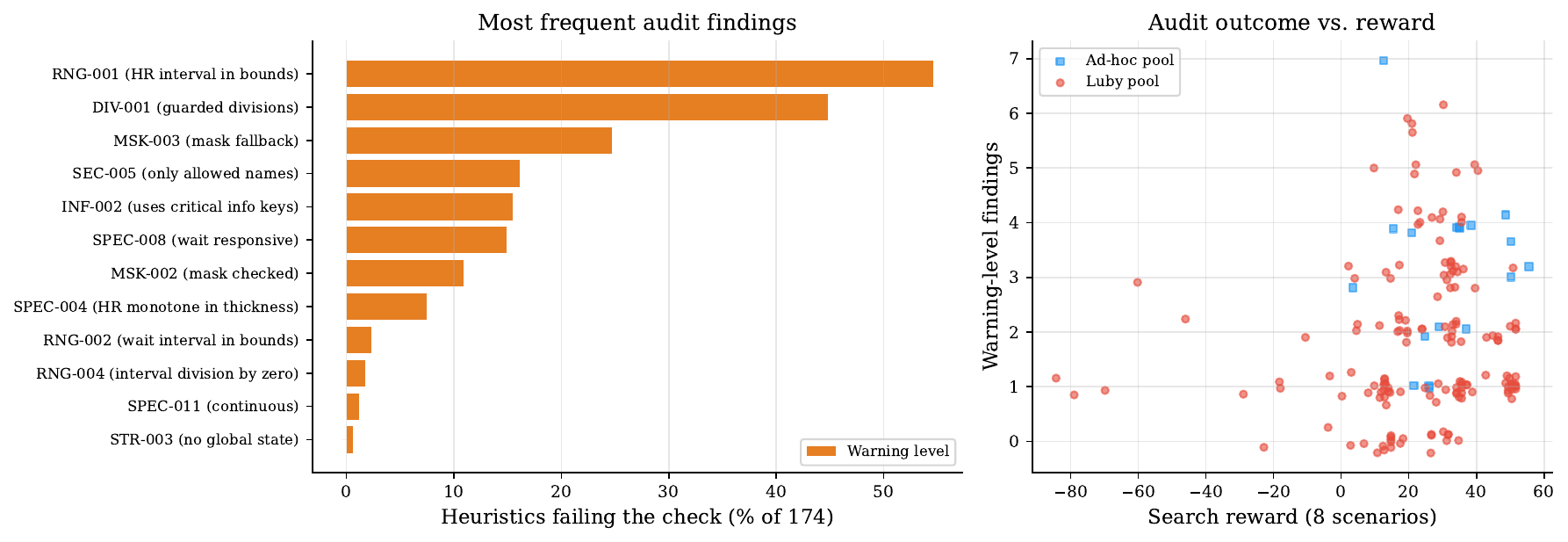}
\caption{Population audit of the 174~synthesized heuristics. Left: share of heuristics failing each of the most frequent checks, all at warning level. Right: warning-level findings against search reward for the ad-hoc (squares) and Luby (dots) pools, with vertical jitter. Warnings increase slightly with reward (Spearman $\rho = 0.16$), so they are not a quality signal.}
\label{fig:audit_population}
\end{figure}

Figure~\ref{fig:audit_population} summarises the population audit of Table~\ref{tab:audit_population}. All findings are warnings: the interval analyser cannot confirm that the HR output stays within bounds for 55\% of the heuristics (RNG-001), 45\% contain unguarded divisions (DIV-001), and 7\% violate HR--thickness monotonicity on the test inputs (SPEC-004). The audit takes a median of 0.05\,s per heuristic and at most 0.24\,s (one AMD EPYC 9654 node, four parallel workers).

Running the audit at population scale also tested the audit tool itself, and we fixed five defects before producing Table~\ref{tab:audit_population}. First, the property tester generated interpass and velocity masks one entry shorter than the environment's (120 and 6 instead of 121 and 7 entries), which falsely flagged 11~heuristics that index the last entry as violating the safety specifications. Second, its masks allowed an interpass time and a rolling velocity of 0, which the environment's mask blocks; 11~other heuristics take the smallest allowed index from the mask and were therefore falsely flagged as returning out-of-range actions. Third, the interval analyser raised an exception when it converted an unbounded interval to an integer, which crashed it on 2~heuristics (Appendix~\ref{app:intervals}). Fourth, the property tester crashed on the hand-coded baseline, which returns its height reduction as a float; it now converts actions to integers as the evaluator does. Fifth, the Z3 translator kept the earlier value of a variable whose reassignment it could not translate, and skipped \texttt{if} statements whose condition it could not translate together with the returns inside them; for the best heuristic, whose height reduction is chosen from the mask array by a helper function, it thus reasoned about a constant height reduction of~0 and proved SPEC-004, although the heuristic requests a 2.0\,mm pass at a thickness of 7.00\,mm and a 0.7\,mm pass at 7.05\,mm (target 5\,mm, HR limit 20\,mm), by the mechanism of the stall of Section~\ref{sec:shift}: the finishing branch is skipped by a small margin and the strain branch takes over. In the same way it had proved SPEC-004 for 63 and refuted it for 45 heuristics; with the translator corrected (Appendix~\ref{app:z3}), all 174 are deferred to property-based testing. The Z3 encoding of SPEC-005 compared a fixed formula with itself instead of the heuristic and proved it for all 174; it now uses the two-translation technique on the interpass output and proves it for 98.

\section{Additional Search Analyses}
\label{app:search_extra}

\begin{figure}[h]
\centering
\includegraphics[width=0.6\linewidth]{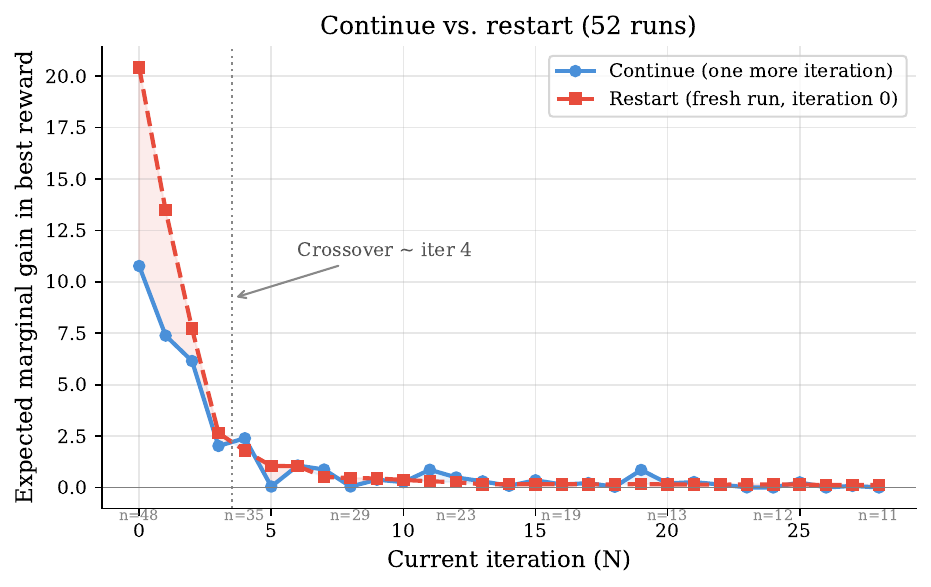}
\caption{Expected marginal gain in best reward from continuing a run for one more iteration versus starting a fresh run, as a function of the current iteration. Restarting pays more during the first few iterations; the curves meet around iteration~4.}
\label{fig:continue_restart}
\end{figure}

Figure~\ref{fig:continue_restart} shows the one-step comparison of continuing a run versus restarting fresh discussed in Section~\ref{sec:budget}. The remaining analyses quantify the uncertainty of the restart results by resampling and simulating the ad-hoc runs. They summarise the global-best distribution of an allocation by its mean rather than by its median as in Figures~\ref{fig:budget} and~\ref{fig:luby_budget}: with a few dozen runs per length, the median of a product of empirical CDFs jumps between adjacent order statistics, which makes resampled and simulated medians erratic.

\begin{figure}[t]
\centering
\begin{minipage}[t]{0.48\linewidth}
\vspace{0pt}\centering
\includegraphics[width=\linewidth]{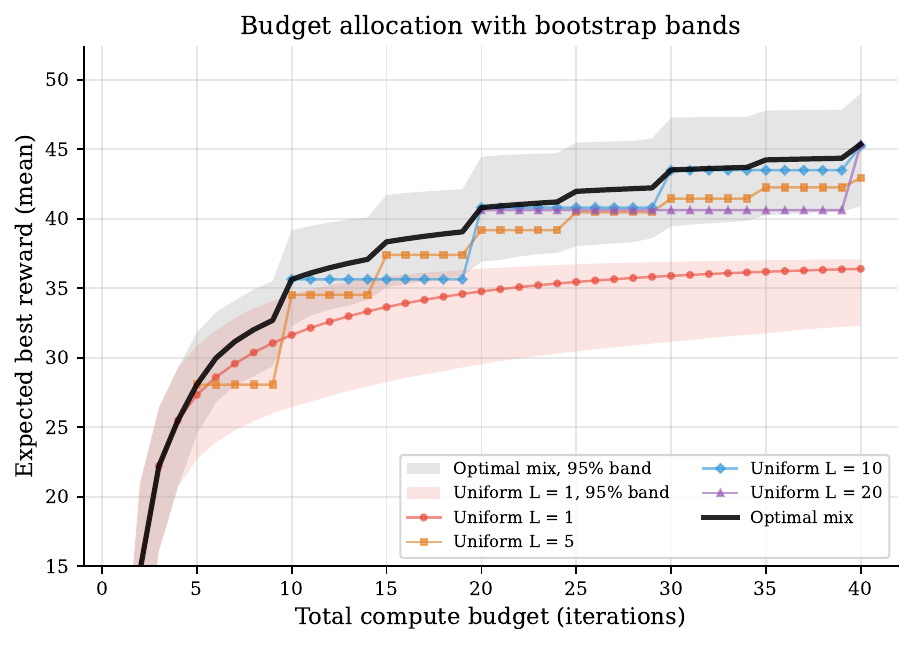}
\caption{Expected best reward (mean) as a function of total iteration budget for the strategies of Figure~\ref{fig:budget}. Shaded: 95\% bootstrap bands for the optimal mix and for one-iteration sampling.}
\label{fig:budget_bootstrap}
\end{minipage}\hfill
\begin{minipage}[t]{0.48\linewidth}
\vspace{0pt}\centering
\includegraphics[width=\linewidth]{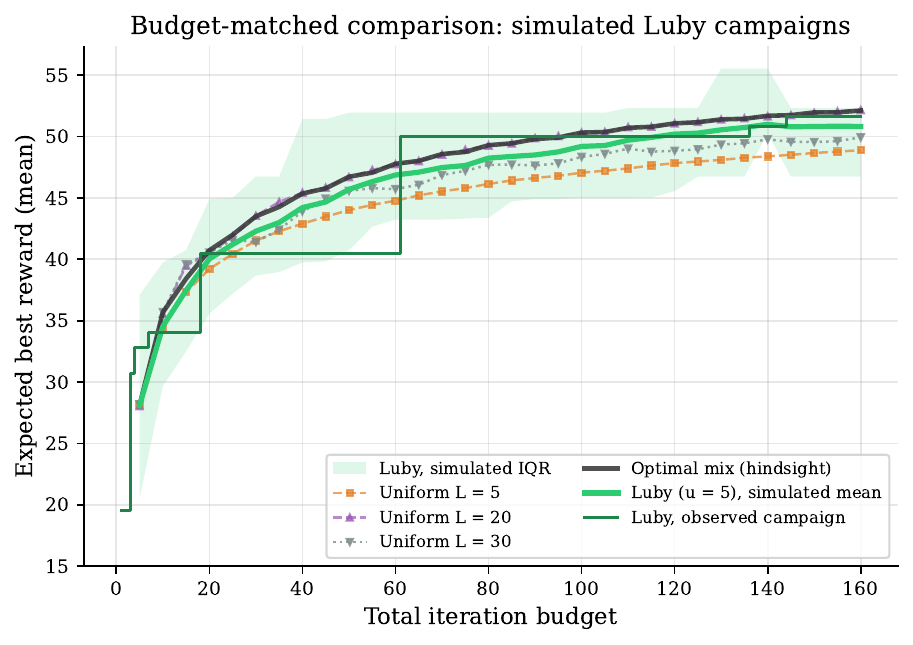}
\caption{Budget-matched comparison by simulation: expected best reward (mean) of fixed allocations and of the Luby schedule ($u{=}5$; mean and interquartile range of 20{,}000 simulated campaigns whose sub-runs are drawn from the per-length distributions of the ad-hoc runs), together with the actual cumulative best of the unseeded Luby campaign, which ends at the 39th percentile of the simulated ones.}
\label{fig:luby_sim}
\end{minipage}
\end{figure}

\paragraph{Bootstrap.} Figure~\ref{fig:budget_bootstrap} repeats the analysis of Figure~\ref{fig:budget} with 95\% confidence bands from resampling runs (1000 bootstrap replicates, with the allocation fixed to the one that is optimal on the full data). At $T{=}20$, the optimal mix (10+10) reaches 40.8 (95\% CI $[36.9, 44.5]$) and a single run of~20 reaches 40.6; the paired bootstrap difference is $+0.1$ (CI $[-2.5, +2.1]$), and the mix is better in only 53\% of resamples. Against four runs of~5, the mix gains $+1.6$ (CI $[-1.1, +5.2]$, better in 85\% of resamples). Pure one-iteration sampling is clearly worse as the budget grows: at $T{=}40$ it reaches 36.4 (CI $[32.3, 37.1]$) against 45.4 (CI $[40.9, 49.0]$) for the optimal mix.

\paragraph{Subsets of runs.} Because runs were stopped at different lengths, each $F_L$ is estimated from a different subset of runs, which assumes that when a run was stopped is unrelated to how well it was doing. The data do not support this assumption: 35~runs ended before their planned length, and for 77\% of them the final best-so-far lies below the median of the runs that continued past the same iteration (at the 28th percentile on average). Table~\ref{tab:restart_subsets} therefore repeats the analysis on three subsets: the 49~runs with Gemini 2.5 Pro, the 27~runs with the final settings (Gemini 2.5 Pro, temperature 0.7 or 1.0, three ideas, stagnation threshold~6; Appendix~\ref{app:configs}), and the 17~runs that reached their planned length, for which stopping cannot depend on performance. In all of them, the optimal allocation at $T{=}20$ is 10+10 and is not reliably better than a single run of 20 or four runs of~5, one-iteration sampling falls clearly behind, and the observed Luby campaign is typical. On the two smaller subsets, the optimal allocations for $T{=}30$ and $T{=}40$ consist of runs of 10 rather than 20~iterations, and runs of 30~iterations fall further behind the optimal allocation (by 2.2--4.1 instead of 0.8--2.4~points).

\begin{table}[t]
\centering
\caption{Restart analysis on subsets of the ad-hoc runs: paired bootstrap differences at $T{=}20$ between the optimal allocation (10+10 in every subset) and two uniform allocations (median and 95\% CI), expected best reward at $T{=}40$ of one-iteration sampling and of the optimal allocation, and the percentile of the observed Luby campaign among 20{,}000 simulated ones.}
\label{tab:restart_subsets}
\small
\begin{tabular}{@{}lrccrrr@{}}
\toprule
 & & \multicolumn{2}{c}{$T{=}20$: optimal allocation minus} & \multicolumn{2}{c}{$T{=}40$} & Luby \\
\cmidrule(lr){3-4}\cmidrule(lr){5-6}
Runs & $n$ & one run of 20 & four runs of 5 & $L{=}1$ & optimal & percentile \\
\midrule
All & 52 & $+0.1$ $[-2.5, +2.1]$ & $+1.6$ $[-1.1, +5.2]$ & 36.4 & 45.4 & 39 \\
Gemini 2.5 Pro & 49 & $+0.1$ $[-2.5, +2.3]$ & $+1.4$ $[-1.1, +5.1]$ & 36.4 & 45.4 & 38 \\
Final settings & 27 & $+1.2$ $[-1.4, +3.6]$ & $+0.8$ $[-2.3, +5.6]$ & 37.0 & 46.8 & 28 \\
Reached planned length & 17 & $+2.1$ $[-0.8, +4.2]$ & $+0.9$ $[-3.2, +6.6]$ & 36.2 & 47.2 & 24 \\
\bottomrule
\end{tabular}
\end{table}

\paragraph{Simulated campaigns.} Figure~\ref{fig:luby_sim} compares the Luby schedule ($u{=}5$) with fixed allocations at matched budgets by simulation. Each sub-run's best reward is drawn from the empirical distribution~$F_L$ of its length; sub-runs of 40 iterations use $F_{30}$, the longest observed, which assumes that Luby sub-runs behave like ad-hoc runs of the same length, and the last run of a fixed allocation is truncated to fit the budget. After 160~iterations, 20{,}000 simulated Luby campaigns reach a mean best reward of 50.9 (median 51.9, interquartile range $[46.7, 52.3]$); the observed campaign's 51.6 lies at the 39th percentile, and 22\% of the campaigns reach 55.5. Uniform runs of 10 or 20~iterations stay within 0.3~points of the hindsight-optimal allocation at every budget, runs of 5~iterations trail it by 1.5--3.3~points from $T{=}20$ on, and runs of 30~iterations by 0.8--2.4~points from $T{=}30$ on.

\begin{figure}[t]
\centering
\includegraphics[width=\linewidth]{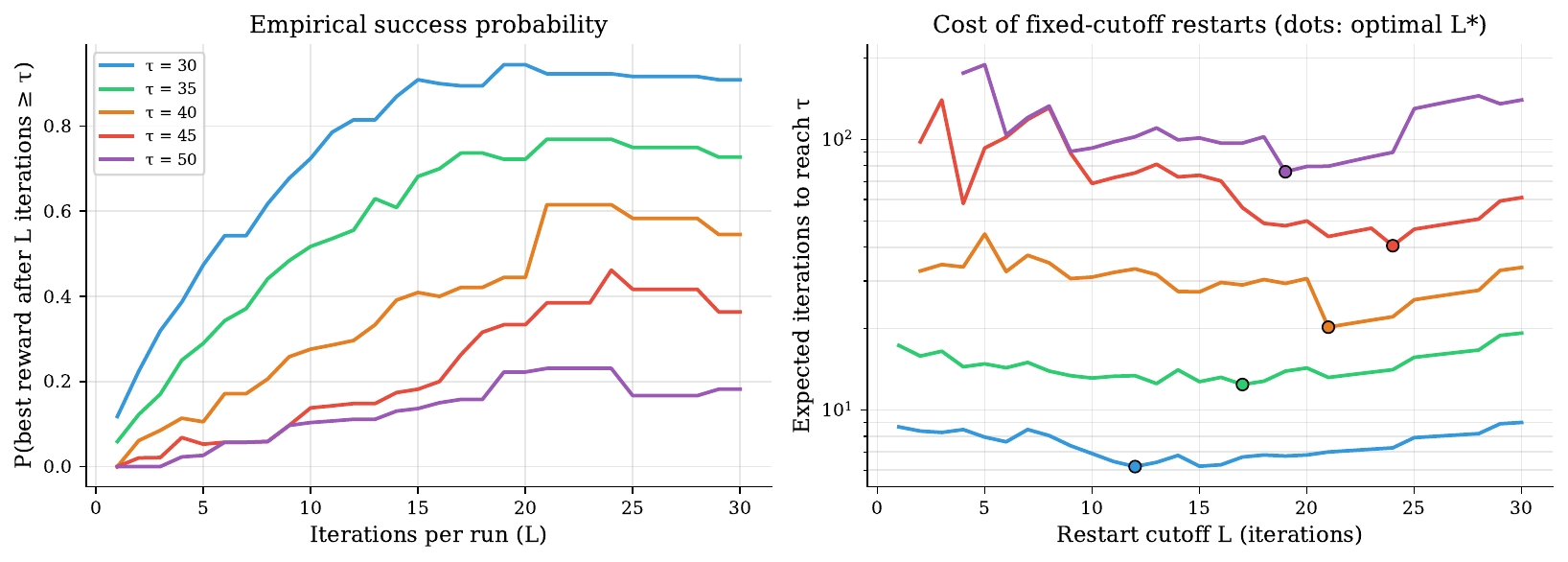}
\caption{Restart analysis at fixed reward thresholds~$\tau$. Left: empirical probability that a run's best-so-far reward after $L$ iterations is at least~$\tau$. Right: expected iterations to reach~$\tau$ when restarting every $L$ iterations, $\mathbb{E}[\min(X, L)]/P(X \leq L)$, in the same colors; dots mark the optimal cutoff $L^*$, which depends on the target.}
\label{fig:threshold}
\end{figure}

\paragraph{Time to a reward threshold.} Figure~\ref{fig:threshold} shows the empirical success curves and fixed-cutoff restart costs behind Table~\ref{tab:threshold}. The success curves are not monotone in $L$ because each length is estimated from the subset of runs that reached it.

\subsection{Run Configurations}
\label{app:configs}

Table~\ref{tab:config} summarises the search configuration of the ad-hoc runs, and Table~\ref{tab:configs} lists the groups of runs with identical settings. All runs use the same 8~evaluation scenarios, reward, and simulator, and three repair attempts per heuristic. Settings marked ``--'' were not recorded in the run logs.

\begin{table}[t]
\centering
\caption{Search configuration of the 52 ad-hoc runs. For settings that changed between groups of runs, the number of runs is given in parentheses (Table~\ref{tab:configs}).}
\label{tab:config}
\small
\begin{tabular}{@{}lr@{}}
\toprule
Parameter & Value \\
\midrule
LLM model & Gemini 2.5 Pro (49), 2.5 Flash (3) \\
Temperature & 0.7 (18), 1.0 (10), 1.5 (2), \\
 & not recorded (22) \\
Outer iterations per run & 1--30 \\
Independent runs & 52 \\
Ideas per iteration & 3 (43), 2 (7), 4 (2) \\
Repair attempts per heuristic & 3 \\
Evaluation scenarios & 8 \\
Max episode length & 25 passes \\
Stagnation threshold & 6 (37), 8 (8), not recorded (7) \\
Conversation memory & compaction (36), synthesis (16) \\
Compaction threshold & max 20--40, keep 10 \\
\bottomrule
\end{tabular}
\end{table}

\begin{table}[t]
\centering
\caption{Groups of ad-hoc runs with identical settings, in order of their first run (all in February 2026). T: sampling temperature; Stag.: stagnation threshold; compaction $a/b$: compact at $a$ messages, keep $b$.}
\label{tab:configs}
\footnotesize
\setlength{\tabcolsep}{3pt}
\begin{tabular}{@{}lrlrrrl@{}}
\toprule
Runs started & $n$ & Model & T & Ideas & Stag. & Memory \\
\midrule
19 21:11--23:14 & 4 & Pro & -- & 2 & -- & compaction 20/10 \\
19 23:23--20 00:03 & 3 & Pro & -- & 3 & -- & compaction 20/10 \\
20 00:58--03:07 & 3 & Pro & -- & 3 & 8 & compaction 20/10 \\
20 01:19--02:21 & 3 & Pro & -- & 2 & 8 & compaction 20/10 \\
20 02:24--02:58 & 2 & Pro & -- & 4 & 8 & compaction 20/10 \\
20 03:52--09:27 & 4 & Pro & -- & 3 & 6 & compaction 25/10 \\
20 10:05--10:29 & 2 & Pro & -- & 3 & 6 & compaction 40/10 \\
20 10:37 & 1 & Flash & -- & 3 & 6 & compaction 25/10 \\
20 10:59 & 1 & Flash & 1.5 & 3 & 6 & compaction 25/10 \\
20 11:02 & 1 & Flash & 1.0 & 3 & 6 & compaction 25/10 \\
20 11:08 & 1 & Pro & 1.0 & 3 & 6 & compaction 25/10 \\
20 11:48 & 1 & Pro & 1.5 & 3 & 6 & compaction 25/10 \\
20 11:53--21 12:20 & 9 & Pro & 0.7 & 3 & 6 & compaction 25/10 \\
21 16:27--22 00:08 & 9 & Pro & 0.7 & 3 & 6 & synthesis \\
22 10:25--23 22:34 & 7 & Pro & 1.0 & 3 & 6 & synthesis \\
22 11:45 & 1 & Pro & 1.0 & 3 & 6 & compaction 20/10 \\
\bottomrule
\end{tabular}
\end{table}

\subsection{Memory Strategy}
\label{app:memory}

\begin{figure}[t]
\centering
\includegraphics[width=0.6\linewidth]{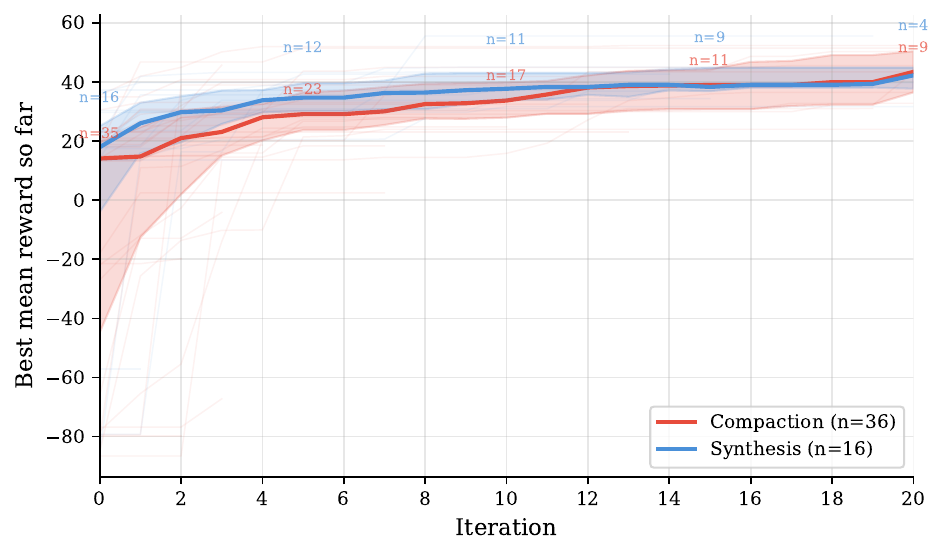}
\caption{Best-so-far reward by memory strategy. \emph{Synthesis} resets the conversation each iteration and provides the LLM with a structured ${\sim}$800-word knowledge summary synthesised by a narrator; \emph{Compaction} preserves the full conversation and summarises old messages when the context grows too large. Solid lines show medians; shaded bands show IQR. Labels give the number of runs contributing at each iteration, which decreases as shorter runs drop out. The synthesis strategy shows higher median reward in early iterations, though sample sizes are modest (16 vs.\ 36 runs) and the two groups were run sequentially, so this comparison should be interpreted cautiously.}
\label{fig:memory_strategy}
\end{figure}

To address conversation amnesia, we compared two memory strategies: \emph{compaction} (the default, which summarises old messages when the context grows too large) and \emph{synthesis} (which resets the conversation each iteration and provides the LLM with a structured knowledge document maintained by a narrator agent). We report this comparison as an exploratory observation. Figure~\ref{fig:memory_strategy} shows that synthesis runs achieve higher median reward over the first ten iterations, with the gap most pronounced in the first 5~iterations; afterwards the medians are close. The best overall heuristic (reward~55.5) came from a synthesis run. However, we caution that sample sizes are modest (16~synthesis vs.\ 36~compaction runs) and the two groups were run sequentially rather than interleaved, so confounds such as prompt refinements between batches cannot be ruled out. More controlled experiments are needed to establish a robust conclusion.

\subsection{Search Trace Excerpt}
\label{app:trace}

Table~\ref{tab:search_trace} is the excerpt of a 20-iteration search discussed in \S\ref{sec:search_interp}.

\begin{table}[t]
\centering
\caption{Strategy evolution over a 20-iteration search, in the LLM's own words. Reward is the mean across 8 scenarios.}
\label{tab:search_trace}
\small
\begin{tabular}{@{}cp{0.72\linewidth}r@{}}
\toprule
Iter & LLM's stated strategy & Reward \\
\midrule
0 & ``Adaptive Two-Phase: bulk reduction then precision finishing'' & 23.5 \\
1 & ``Fix catastrophic temperature failure: \emph{every} scenario loses ${\sim}$2.0 pts'' & 23.3 \\
3 & ``Proactive heat conservation via interpass time modulation'' & 33.8 \\
9 & ``Temperature excels but core reduction strategy has systemic flaws'' & 33.1 \\
10 & ``Radical departure: the conflict between thermal and metallurgical efficiency requires a fundamentally new approach'' & 16.6 \\
14 & ``Fuse strain-based HR with re-tuned thermal response---hybrid synthesis'' & \textbf{38.0} \\
\bottomrule
\end{tabular}
\end{table}

\section{Out-of-Distribution Scenarios}
\label{app:shift}

\begin{table}[t]
\centering
\caption{Mean episode return per out-of-distribution family (ad-hoc pool / Luby pool). The best single heuristic and the $K{=}4$ portfolio are selected on the in-distribution grid; the pool oracle keeps the best heuristic of the whole pool per case.}
\label{tab:shift_families}
\small
\begin{tabular}{@{}lrrrr@{}}
\toprule
Family (cases) & Baseline & Best single & Portfolio $K{=}4$ & Pool oracle \\
\midrule
Thicker slabs, 130--150\,mm (6) & 38.6 & 56.9 / 60.5 & 62.4 / 68.1 & 63.9 / 68.2 \\
Thin targets, 2--4\,mm (6) & 33.1 & $-$17.2 / 43.0 & 49.1 / 50.4 & 49.1 / 62.8 \\
Thick targets, 20--40\,mm (6) & 20.8 & 27.9 / 28.3 & 38.7 / 31.9 & 46.0 / 47.0 \\
Tight mills, HR limit 10--15\,mm (6) & 32.5 & 84.5 / 79.3 & 91.4 / 85.6 & 91.9 / 91.8 \\
Loose mills, HR limit 55--60\,mm (4) & 13.2 & 34.7 / 8.7 & 34.8 / 39.1 & 35.1 / 40.3 \\
Off-grid temperature pairs (4) & 40.9 & 62.8 / 53.4 & 64.5 / 63.1 & 66.1 / 64.1 \\
Off-grid grain size pairs (4) & 60.2 & 73.1 / 61.0 & 73.9 / 69.3 & 75.7 / 71.5 \\
Combined shifts (6) & 59.5 & 26.9 / 65.1 & 53.6 / 69.0 & 76.4 / 84.0 \\
\midrule
All 42 cases & 37.2 & 41.8 / 51.2 & 58.7 / 59.9 & 63.6 / 67.3 \\
\bottomrule
\end{tabular}
\end{table}

\begin{figure}[t]
\centering
\includegraphics[width=0.6\linewidth]{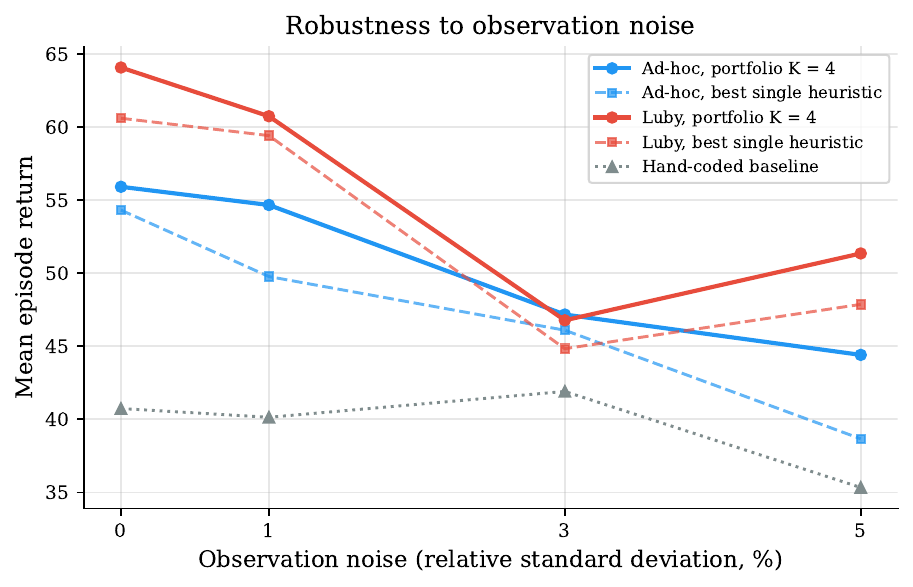}
\caption{Mean episode return under multiplicative Gaussian observation noise for the best single heuristic and the $K{=}4$ portfolio of each pool, both selected on the in-distribution grid, and for the hand-coded baseline (three scenarios $\times$ three seeds per noise level; the noise-free values are for the same three scenarios).}
\label{fig:noise}
\end{figure}

All eight families start from a nominal scenario (100$\to$10\,mm, grain size 200$\to$12\,$\mu$m, temperature 1373$\to$1173\,K, HR limit 30\,mm) and change it as follows: (1)~thicker slabs of 130, 140, and 150\,mm with targets of 8 and 12\,mm; (2)~thin targets of 2, 3, and 4\,mm from 80 and 100\,mm; (3)~thick targets of 20, 30, and 40\,mm from 100 and 120\,mm; (4)~tight mills with HR limits of 10, 12, and 15\,mm from 90 and 110\,mm; (5)~loose mills with HR limits of 55 and 60\,mm from 100 and 120\,mm; (6)~temperature pairs (initial$\to$target) 1300$\to$1250, 1500$\to$1100, 1450$\to$1080, and 1290$\to$1200\,K; (7)~grain size pairs 120$\to$6, 480$\to$24, 150$\to$22, and 450$\to$7\,$\mu$m; and (8)~six combined shifts that change all seven scenario parameters at once (e.g., 140$\to$4\,mm, 450$\to$8\,$\mu$m, 1500$\to$1100\,K, HR limit 22\,mm). Table~\ref{tab:shift_families} reports the results per family. The observation-noise experiment uses three scenarios of the in-distribution grid (100$\to$10, 110$\to$15, and 90$\to$5\,mm at nominal grain size, temperature, and HR limit) with three seeds per noise level; the noise is applied to the observation the heuristic sees, not to the simulator state. Figure~\ref{fig:noise} shows the results; with nine episodes per noise level, differences of a few points are within noise.

Five episodes in which the heuristics of one Luby sub-run raise an exception (they index the 501-entry HR array beyond its end when the HR limit is 60\,mm, above the 50\,mm action range) are excluded.

\end{document}